\documentclass[11pt]{article}

\usepackage[preprint]{acl}

\usepackage{times}
\usepackage{latexsym}

\usepackage[T1]{fontenc}
\usepackage[utf8]{inputenc}
\usepackage{bbm}

\usepackage{microtype}

\usepackage{inconsolata}

\usepackage{graphicx}

\usepackage{hyperref}
\usepackage{url}
\usepackage{algorithm}
\usepackage{algorithmic}
\usepackage{subfiles}
\usepackage{booktabs}       % professional-quality tables
\usepackage{amsmath,mathtools}
\usepackage{amssymb}
\usepackage{amsfonts}

\usepackage{multirow}
\usepackage{multicol}
\usepackage{xcolor}
\usepackage{adjustbox}
\usepackage{amsthm}
\theoremstyle{definition}
\newtheorem{definition}{Definition}

\theoremstyle{remark}

\usepackage{pifont}
\usepackage{caption}
\usepackage{subcaption}
\usepackage{wrapfig}
\usepackage{simplebnf}
\usepackage{tcolorbox}
\usepackage{tikz}
\usepackage{tikz-qtree}
\usepackage{dashrule}
\tcbuselibrary{listings,breakable}
\tcbset{listing engine=listings,colframe=black,colback=white,size=small}
\usepackage{enumitem}

\usepackage{booktabs,multirow,tabularx,makecell,array}
\newcolumntype{Y}{>{\raggedright\arraybackslash}X} % wrapping text column

\usepackage[table]{xcolor}

\usepackage{graphicx}
\usepackage{svg} % only needed if using .svg files

\usepackage{fvextra}
\DefineVerbatimEnvironment{Prompt}{Verbatim}{
  fontsize=\footnotesize,
  breaklines=true,
  breakanywhere=true
}

\definecolor{lightgreen}{RGB}{220,245,220}
\definecolor{lightred}{RGB}{255,220,220}

\definecolor{LightYellow}{RGB}{255,250,208}
\definecolor{LightGreen}{RGB}{215, 245, 203}
\definecolor{LightBlue}{RGB}{208, 245, 243}
\definecolor{LightPurple}{RGB}{224,223,255}
\definecolor{LightGrey}{RGB}{225,225,225}

\usepackage{soul}
\usepackage{needspace}
\tcbuselibrary{skins}
\definecolor{qexprompt}{HTML}{888780}
\definecolor{qexours}{HTML}{185FA5}
\definecolor{qexrar}{HTML}{5F5E5A}

\newtcolorbox{qexbox}[2]{enhanced, breakable, beforeafter skip balanced=0.5\baselineskip,
  colframe=#1, colback=white,
  colbacktitle=#1!10!white, coltitle=#1!70!black, boxrule=0.5pt, arc=1pt,
  left=6pt, right=6pt, top=4pt, bottom=4pt, toptitle=2pt, bottomtitle=2pt,
  fontupper=\scriptsize, fonttitle=\small, title={\textbf{#2}},
  before upper={\setlength{\parskip}{1pt}\setlength{\parindent}{0pt}}}
\title{PaperCritic: Training AI Reviewers to Assess the Soundness of Papers}

\author{
 \textbf{Hyun Ryu\textsuperscript{1,2}},
 \textbf{Ji Yong Cho\textsuperscript{3}},
 \textbf{Eunho Yang\textsuperscript{1}},
 \textbf{Moontae Lee\textsuperscript{3}},
\\
 \textbf{June Yong Yang\textsuperscript{3}}\thanks{Corresponding authors.},
 \textbf{Sean Welleck\textsuperscript{2}}\footnotemark[1],
 \textbf{Carolyn Ros\'e\textsuperscript{2}}\footnotemark[1]
\\
\\
 \textsuperscript{1}KAIST AI,
 \textsuperscript{2}Carnegie Mellon University Language Technologies Institute,
 \textsuperscript{3}LG AI Research
\\
 \texttt{\{ryuhyun1905,eunhoy\}@kaist.ac.kr,~~\{swelleck,cprose\}@cs.cmu.edu,} \\
 \texttt{\{jiyong.cho,moontae.lee,juneyong.yang\}@lgresearch.ai}
}

\begin{document}
\maketitle
\begin{abstract}
% \vspace{-3mm}
As AI conference submissions grow rapidly, AI reviewers are being considered as an alternative.
However, simply imitating human reviewing is problematic because reviewing serves multiple purposes, including evaluating papers' soundness, interestingness, and clarity, the last two of which are subjective~\citep{tmlr}.
In this work, we focus solely on evaluating soundness, the most important and objective criterion, and define the learning objective as jointly maximizing (1) precision, based on the \textit{soundness} and \textit{substantiveness} of review points, and (2) recall, the coverage of high-quality reference review points.
Under this problem formulation, we construct a quality-labeled review point dataset, \texttt{PaperSound-23K}, using 1K ICLR submissions and 23K human-written and LLM-generated review points.
Using this dataset, we train an AI reviewer model, PaperCritic, through reinforcement learning with the F1 score of precision and recall as the reward.
Our experiments show that PaperCritic outperforms frontier models and models trained with other AI reviewer training methods on an in-domain test set.
Surprisingly, learning to review ICLR submissions also transfers to reviewing Nature-family papers and even to out-of-domain reviewing tasks (e.g., code review, legal contract review, and feedback on story writing).
% \footnote{The source code and data will be publicly available.}
% \vspace{-3mm}
\end{abstract}

%%%%%%%
\section{Introduction}
\label{sec:intro}

As AI conference submissions grow exponentially, the demand for reviews grows at the same rate~\citep{position, reviewscore}. Since the number of qualified human reviewers is limited, large language models (LLMs) are being considered as complementary paper reviewing assistants.
Prior work proposes several AI reviewer training methods~\citep{openreviewer, deepreview, cyclereview, reviewrl, guide}. These studies formulate paper reviewing as \textit{holistic review-and-score generation} to (1) maximize review quality relative to a reference human review and/or (2) minimize the discrepancy between generated and human overall scores. This is essentially a human review imitation problem.

\begin{figure}[t]
  \vspace{-8mm}
  \centering
  \includegraphics[width=0.87\columnwidth]{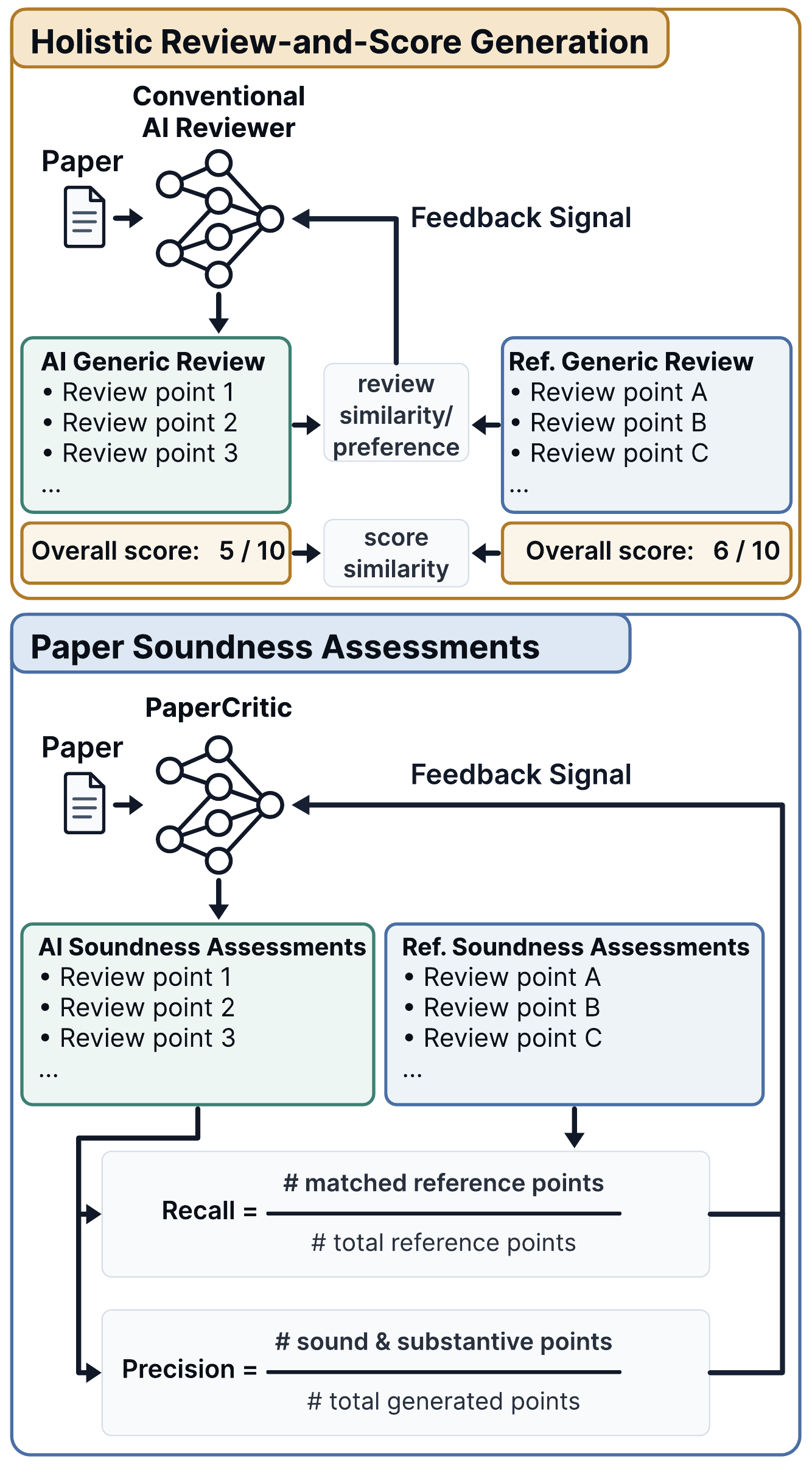}
  \caption{
  \textbf{Comparison of paper soundness assessments with conventional holistic review-and-score generation.}
  Conventional AI reviewers are trained using holistic review similarity or preference and score similarity as objectives.
  In contrast, our learning objective is to jointly maximize (1) the \textit{soundness} and \textit{substantiveness} of review points and (2) the \textit{coverage} of sound and substantive reference review points.
  }
  \label{fig:problem_formulation}
  \vspace{-6mm}
\end{figure}

Training AI reviewers beyond human review imitation requires understanding the fundamental purpose of paper reviewing and defining an objective function that captures it.
According to \citet{tmlr}, there are three main purposes of paper reviewing: evaluating papers' soundness, interestingness, and clarity. Here, soundness refers to whether the claims made in a paper are supported by accurate and convincing evidence and is the most important criterion. The other two criteria are more subjective and therefore need to be treated carefully.
In this work, we focus solely on evaluating the soundness of papers with AI reviewers.

To identify an objective function for the paper soundness assessment task, we conduct a human study\footnote{The detailed procedure for the human study is described in Appendix~\ref{app:subsec:human-objectives}.} with seven AI researchers and identify two universal evaluation criteria (i.e., objectives): \textit{soundness}\footnote{In this work, \textit{soundness} refers to the soundness of either papers or their assessments. To distinguish the two, we use \textit{paper soundness} for the former.} and \textit{substantiveness}, formally defined as follows.
\begin{definition}[Sound and Substantive Review Point]
\label{def:sound-substantive}
A review point\footnote{Following \citet{reviewscore}, we call a single, self-contained unit of assessment a \textit{review point} and a list of review points a \textit{review}.} is:
\begin{itemize}[leftmargin=5mm]
\item \textbf{Sound} if and only if it relies on correct premises according to the paper or knowledge in the field.
\vspace{-5mm}
\item \textbf{Substantive} if and only if it raises a moderate to significant concern that a main claim of the paper or its central evidence may not hold.
\end{itemize}
\end{definition}
We also add another objective, \textit{completeness}, which measures the coverage of high-quality (i.e., sound and substantive) reference assessments.
We treat soundness and substantiveness as the precision term and completeness as the recall term, and formulate paper soundness assessment as jointly maximizing precision and recall (Figure~\ref{fig:problem_formulation}).

To train AI reviewers to assess the soundness of papers, we construct a quality-labeled review point dataset named \texttt{PaperSound-23K}.
We sample 1K ICLR submissions and collect 23K human-written and LLM-generated review points.
To evaluate the quality (i.e., soundness and substantiveness) of these review points, we automate the process using an LLM with strong reasoning ability and observe human-model agreement comparable to human-human agreement (Table~\ref{tab:dataset_quality}).

We train PaperCritic on \texttt{PaperSound-23K} training set through supervised fine-tuning (SFT) on sound and substantive reference points, followed by online reinforcement learning (RL) with an F1 reward.
Two reward models (RMs) assess soundness and substantiveness to estimate precision, while a matcher measures reference coverage to estimate recall.

We demonstrate the effectiveness of PaperCritic on the in-domain \texttt{PaperSound-23K} test set and four out-of-domain (OOD) benchmarks covering Nature-family paper review, code review, legal contract review, and feedback on story writing.
PaperCritic-4B outperforms other 4B models trained with conventional methods and frontier models on the in-domain dataset, and our RL training on ICLR paper reviewing consistently improves the model's performance on all four benchmarks.
We further show that our training method remains effective when using a different LLM grader for evaluation and when training a larger model.

Our key contributions are:
\begin{enumerate}[leftmargin=7mm]
\vspace{-2mm}
    \item We introduce the task of paper soundness assessment and formalize it as maximizing two objectives: (1) precision based on the soundness and substantiveness of review points and (2) coverage of reference points.
    \item We build \texttt{PaperSound-23K}, a large-scale dataset for training AI reviewers to assess the soundness of papers.
    \item We propose a simple training recipe including RL with the F1 score of precision and recall as the reward that produces PaperCritic, which outperforms models trained with conventional methods and frontier models on the in-domain test set and effectively generalizes to four out-of-domain benchmarks.
\vspace{-2mm}
\end{enumerate}

\section{Related Work}
\label{sec:related-works}

\paragraph{Paper Review Datasets.}
Several studies have constructed paper review datasets by collecting paper drafts and expert-written reviews from different venues.
Some early studies~\cite{peerread, nlpeer} target preliminary tasks such as paper acceptance classification or review score prediction but do not cover full review generation.
In contrast, more recent datasets~\cite{reviewer2, cyclereview, deepreview} are constructed specifically for the review generation task.
Two important considerations are that reviewers have varied purposes when writing reviews and that the multiple points within a review can vary in quality. However, the aforementioned datasets do not account for these two factors and primarily focus on collecting reviews at scale for model training.
To address these issues, we narrow the scope of paper reviewing to paper soundness assessment and construct \texttt{PaperSound-23K} by labeling the quality of collected review points according to Definition~\ref{def:sound-substantive}.

\paragraph{AI Reviewer Training.}
Another line of work explicitly trains specialized reviewer models.
\citet{openreviewer}, \citet{cyclereview}, and \citet{deepreview} train generic reviewing models by applying SFT to datasets of human-written or LLM-synthesized reviews.
\citet{reviewrl} and \citet{guide} also train generic reviewing models, but apply online RL to further improve review quality. These studies use holistic review similarity or preference relative to human reviews, together with overall score similarity, as rewards for model training.
However, these methods provide sparse supervision and remain limited to imitating human reviews. Our work addresses these limitations by using evaluations of individual review points to provide dense rewards and defining a new training objective.
In addition, \citet{goodpoint} trains constructive feedback generators by applying SFT and DPO using author responses as quality labels for reviews. However, author responses provide noisy labels of review quality~\citep{goldberg} and are difficult to use for online reward computation. We instead identify reliable review quality criteria under our problem formulation and use corresponding reward models that enable online learning.

\section{Problem Formulation}
\label{sec:problem-formulation}

\paragraph{Prior formulation: holistic review-and-score generation.}
Given a paper and optional context, prior methods generate a holistic review and an overall score. Training uses SFT on human or synthesized reviews~\citep{openreviewer,deepreview,cyclereview}, review-similarity rewards~\citep{guide}, or preference-based rewards~\citep{reviewrl}, together with alignment to reference scores.

\paragraph{Our formulation: paper soundness assessments.}
In contrast, we formulate paper reviewing as generating a list of individual review points that identify meaningful concerns about the soundness of a paper.
Given a target paper \(x\), the model generates $Y = \{y_1, y_2, \ldots, y_m\}$.
A review point is considered \emph{high-quality} when it is both \textit{sound} and \textit{substantive} (Definition~\ref{def:sound-substantive}):
\[
Q(x,y_i)
=
\mathrm{Sound}(x,y_i)
\cdot
\mathrm{Substantive}(x,y_i)
\]
where $\mathrm{Sound}(x,y_i) \in \{0,1\}$ indicates whether $y_i$ is sound and $\mathrm{Substantive}(x,y_i) \in \{0,1\}$ indicates whether $y_i$ is substantive.

Our objective has two complementary components.
First, the generated review should have high \emph{precision}: most generated review points should themselves be sound and substantive. We define
\[
\mathrm{Prec}(x,Y)
=
\frac{1}{|Y|}
\sum_{y_i \in Y}
Q(x,y_i).
\]
Second, the review should have high \emph{recall}: it should cover as many distinct high-quality concerns as possible.
Let $R^{\star}(x) = \{r^{\star}_1,\ldots,r^{\star}_n\}$ denote the set of sound and substantive reference review points for paper \(x\), and let $M(x,y_i,r^{\star}_j) \in \{0,1\}$ indicate whether a generated review point \(y_i\) and a reference review point \(r^{\star}_j\) express the same underlying concern.
We allow many-to-many matching between generated and reference review points and therefore define recall as the fraction of unique reference review points that are matched by at least one generated review point:
\[
\mathrm{Rec}(x,Y)
=
\frac{1}{|R^{\star}(x)|}
\sum_{j=1}^{|R^{\star}(x)|}
\max_{y_i \in Y}
M(x,y_i,r^{\star}_j).
\]
We therefore formulate review generation as a multi-objective learning problem:
\[
\max_{\theta}\;
\mathbb{E}_{x\sim\mathcal{D},\,Y\sim p_\theta(\cdot\mid x)}
\left[
\big(
\mathrm{Prec}(x,Y),\,
\mathrm{Rec}(x,Y)
\big)
\right],
\]
where both objectives are to be maximized jointly.

\section{\texttt{PaperSound-23K}: A Quality-Labeled Review Point Dataset}
\label{sec:dataset}

To train an AI reviewer under our problem formulation, we build a large-scale dataset named \texttt{PaperSound-23K}, which labels the \textit{soundness} and \textit{substantiveness} of 23K review points for 1K papers.
We describe automated LLM-based dataset construction (Section~\ref{subsec:dataset-construction}), assess label quality through human verification (Section~\ref{subsec:dataset-quality}), and summarize the dataset (Section~\ref{subsec:dataset-analysis}).

\subsection{Dataset Construction}
\label{subsec:dataset-construction}

We describe an end-to-end automated pipeline to build \texttt{PaperSound-23K}.

\paragraph{Step 0: Venue and Paper Selection.}
To obtain initial paper submissions and their corresponding reviews, we use ICLR 2020--2023 submissions on OpenReview as the source.
Then, we select 1K submissions using proportionate stratified sampling across three dimensions: year, acceptance status, and research area\footnote{Details are provided in Appendix~\ref{app:sec:add-stats}.}.

\paragraph{Step 1: Paper and Review Collection.}
We use the OpenReview API to collect the 1K sampled papers and their corresponding human-written reviews, and DeepSeek-OCR~\cite{deepseek-ocr} to convert the papers from PDF to text.
To cover concerns that human reviewers might miss, we also add LLM-generated synthetic reviews.

\paragraph{Step 2: Review Point Extraction.}
An LLM extracts review points from human-written reviews. To do this, we iteratively tune the prompt on a small subset of papers until a human expert verifies that all extractions are correct.
Since the LLM reviewer outputs structured JSON, a regex parser extracts review points from the synthetic reviews.
Finally, we filter out review points that do not assess the paper's soundness or that contain external citations.

\paragraph{Step 3: Claim and Evidence Extraction.}
An LLM extracts 3--6 main claims per paper and 1--4 pieces of evidence per main claim, and a human expert verifies the quality of the extractions on the same subset of papers used in Step 2.

\begin{table}[t]
\centering
\small
\setlength{\tabcolsep}{4pt}
\begin{tabular}{lrr}
\toprule
 & \textbf{Human-Model} & \textbf{Human-Human} \\
\midrule
Soundness     & 0.654 & 0.640 \\
Claim Linkage & 0.611 & 0.606 \\
Severity      & 0.563 & 0.557 \\
\midrule
High-quality (binary)  & 0.624 & 0.597 \\
\bottomrule
\end{tabular}
\caption{
\textbf{Human-model agreement on review point evaluation.}
We measure human-model agreement using Spearman's correlation of soundness, claim linkage, and severity scores and binary aggregated high-quality labels, and compare it with human-human agreement.
Since three human experts annotate the same instances, we report the average of the three human-model correlations and the average of the three pairwise human-human correlations.
}
\label{tab:dataset_quality}
\end{table}

\begin{table}[t]
\centering
\small
\setlength{\tabcolsep}{4pt}
\begin{tabular}{lrrr}
\toprule
 & \textbf{Train} & \textbf{Test} & \textbf{Total} \\
\midrule
\# Papers & 889 & 100 & 989 \\
\# Review points & 20,933 & 2,282 & 23,215 \\
Avg. review points / paper & 23.55 & 22.82 & 23.47 \\
Std. review points / paper & 7.04 & 6.12 & 6.95 \\
\% Human review points & 62.6\% & 61.2\% & 62.4\% \\
\midrule
\multicolumn{4}{c}{\textit{Statistics of Quality Labels}} \\
\addlinespace[2pt]
\text{Sound and substantive}     & 5,674 & 612 & 6,286 \\
\text{Sound but not substantive} & 5,833 & 650 & 6,483 \\
\text{Not sound}                 & 9,426 & 1,020 & 10,446 \\
\bottomrule
\end{tabular}
\caption{
\textbf{Statistics of \texttt{PaperSound-23K}.}
We divide the dataset into training and test sets while preserving the proportions of papers by year, acceptance status, and research area.
}
\label{tab:dataset_statistics}
\vspace{-5mm}
\end{table}

\paragraph{Step 4: Review Point Evaluation.}
An LLM with strong reasoning ability evaluates the soundness and substantiveness of the collected review points.
An LLM evaluator first measures soundness on a 3-point scale\footnote{0: unsound, 1: uncertain, 2: sound}.
At the same time, we divide substantiveness into \textit{claim linkage} (how directly the review point challenges the paper's main claim or evidence) and \textit{severity} (how seriously the review point undermines that main claim).
An LLM evaluator identifies the target claim of each review point and measures claim linkage\footnote{0: unrelated, 1: indirectly related, 2: clearly related, 3: directly challenges} and severity\footnote{0: negligible, 1: minor, 2: moderate, 3: significant} on 4-point scales.
Following Definition~\ref{def:sound-substantive}, given a paper $x$ and a review point $y_i$, $\mathrm{Sound}(x,y_i)=1$ if and only if the soundness score is 2, and $\mathrm{Substantive}(x,y_i)=1$ if and only if both the claim linkage and severity scores are at least 2.

\paragraph{Model Choices.}
We use GPT-5.2/5.4~\citep{gpt5} across all steps and vary the reasoning effort according to task difficulty.
The LLM reviewer in Step 1 is GPT-5.2 (xhigh), the review point extractor in Step 2 is GPT-5.2 (none), the postprocessing model in Step 2 is GPT-5.2 (low), the claim/evidence extractor in Step 3 is GPT-5.2 (low), and the review point evaluator in Step 4 is GPT-5.4 (high).
The full prompts used in each step are provided in Appendix~\ref{app:subsec:dataset-prompts}.

\begin{figure*}[t]
  \vspace{-8mm}
  \centering
  \includegraphics[width=1.0\textwidth]{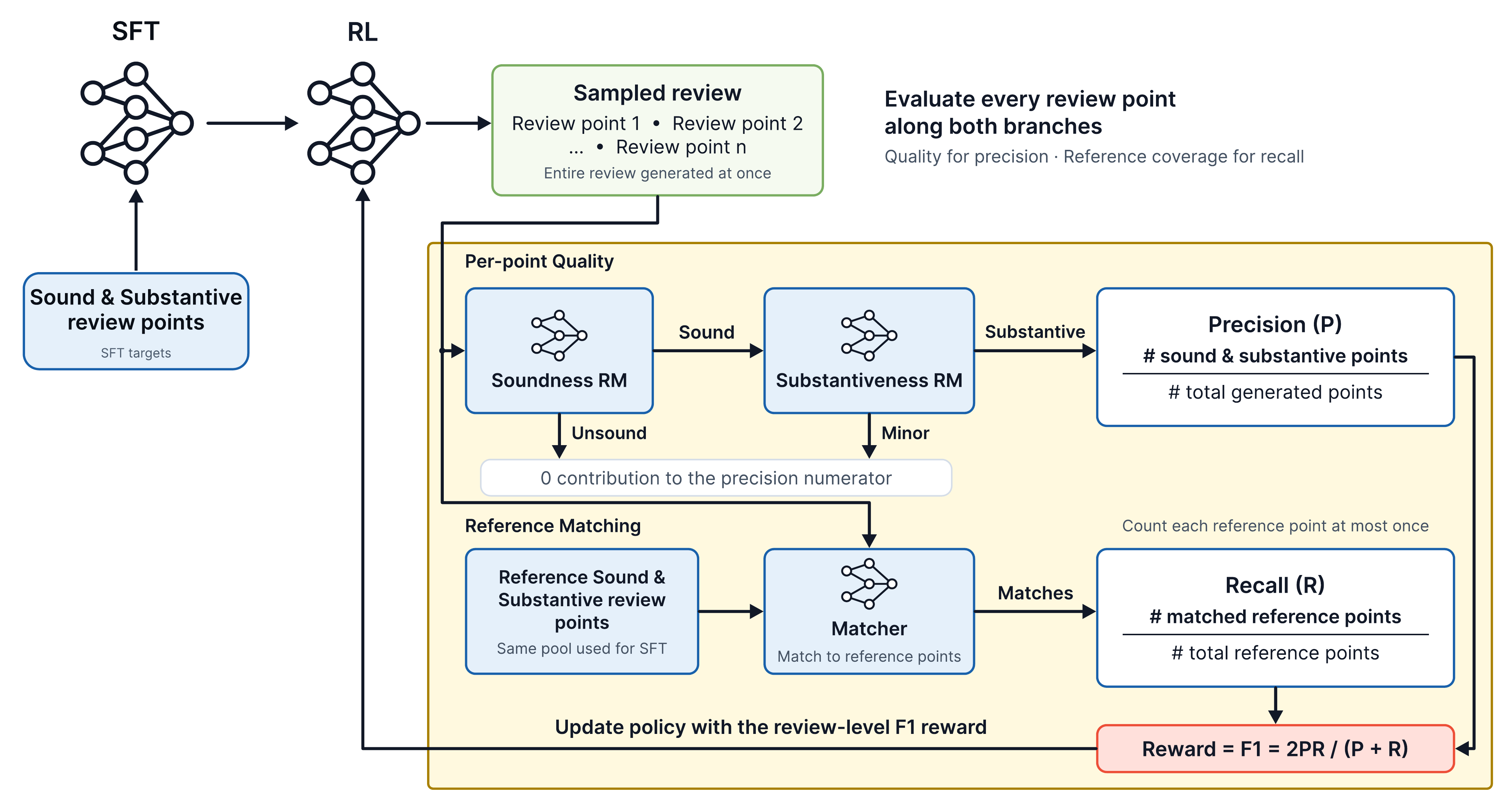}
  \caption{
  \textbf{An overview of the PaperCritic training framework.}
  SFT uses sound and substantive reference points. Online RL maximizes a review-level F1 reward, combining precision estimated by soundness and substantiveness RMs with recall estimated by matching generated and reference points. The RMs use the scoring scales in Section~\ref{subsec:dataset-construction}.
  }
  \label{fig:training_overview}
  \vspace{-5mm}
\end{figure*}

\subsection{Dataset Quality}
\label{subsec:dataset-quality}

We verify the reliability of automated review point evaluation.
To do this, we measure human-model agreement and compare it with human-human agreement (Table~\ref{tab:dataset_quality}).
We recruit three human experts and select seven ICLR submissions from our dataset with a total of 226 review points. The experts then independently annotate the quality (i.e., soundness, claim linkage, and severity) of the same review points\footnote{The detailed human evaluation procedure is described in Appendix~\ref{app:subsec:human-review-quality}.}.
The results show that human-model agreement (0.624) is comparable to human-human agreement (0.597), supporting the effectiveness of LLM-based annotation.

\subsection{Dataset Statistics}
\label{subsec:dataset-analysis}

Table~\ref{tab:dataset_statistics} summarizes \texttt{PaperSound-23K}: 989 papers (889 training, 100 test) and 23K quality-labeled review points (23.47$\pm$6.95 per paper).
Human-written points account for 62.4\% of the data, and LLM-generated points for 37.6\%.
Overall, 27.1\% are sound and substantive, 27.9\% sound but not substantive, and 45.0\% not sound.
Appendix~\ref{app:sec:add-stats} provides further statistics and analysis.

\section{PaperCritic Training Framework}
\label{sec:papercritic}

We train PaperCritic on \texttt{PaperSound-23K} with SFT on sound and substantive reference points (Section~\ref{subsec:sft}), followed by online RL using a reward that combines precision and recall (Section~\ref{subsec:rl}; Figure~\ref{fig:training_overview}).

\subsection{Supervised Fine-Tuning}
\label{subsec:sft}
We train PaperCritic with SFT on sound and substantive review points.
Since review points are collected as an unordered set for each paper, we specify their order to construct training targets.
To do this, we impose a claim-diversity-aware order. We first group sound and substantive review points by their target claims and then interleave points across different claims whenever possible\footnote{If Claim 1 is targeted by \(P_1,P_2,P_3\), Claim 2 by \(P_4\), and Claim 3 by \(P_5,P_6\), one possible order is \([P_1, P_4, P_5, P_2, P_6, P_3]\).}. This encourages the model to cover diverse claims before generating multiple points about the same claim.

\subsection{Online Reinforcement Learning}
\label{subsec:rl}
Following the learning objective for paper soundness assessment (Section~\ref{sec:problem-formulation}), we introduce an online RL training method by specifying (1) how to estimate precision and recall and (2) how to combine them into a single reward.
Given a paper, the policy generates a complete review as a list of review points. We evaluate individual points rather than scoring the review holistically.

\textbf{Precision Estimation.}
We use separate RMs for soundness and substantiveness, evaluating each criterion on the corresponding scoring scale described in Section~\ref{subsec:dataset-construction}.
The soundness RM receives the paper and the generated review point and examines whether the factual premises underlying the review point are supported by the paper or applicable domain knowledge. Only points whose material premises are supported (i.e., soundness score 2) are considered \textit{sound}.
For substantiveness, we provide the paper and its pre-extracted target claims with supporting evidence: a routing stage selects the relevant claim, after which the substantiveness RM evaluates claim linkage and severity scores relative to that claim on 4-point scales. A review point is \textit{substantive} when both scores are at least 2.
Review points that are both sound and substantive contribute to the precision numerator.

\textbf{Recall Estimation.}
An LLM matcher compares each generated point with the reference sound and substantive points (from \texttt{PaperSound-23K}) and returns matched reference points.
We follow the definition of matching in \citet{marg}, which requires both high semantic relatedness and sufficient specificity: the generated point must express essentially the same criticism and be at least as specific as the reference.
Each reference contributes at most once to recall.

\textbf{F1 Reward.}
We combine the estimated precision and recall into an F1 score and use it as a review-level reward.
The harmonic mean encourages a balance between generating sound, substantive criticism and covering reference concerns: high precision alone cannot compensate for low recall, or vice versa. This scalar is assigned to the complete generated review and used as a reward to update the current policy using GRPO~\citep{grpo}.

\section{Experiments}
\label{sec:exp}

Our experiments address four research questions:
\textbf{(RQ1)} How does PaperCritic compare with conventional reviewer-training methods and frontier models in automated and human evaluations?
\textbf{(RQ2)} Does training on ICLR submissions generalize to other scientific domains and reviewing tasks?
\textbf{(RQ3)} How do individual reward components and reward design choices affect performance?
\textbf{(RQ4)} How robust are the observed gains across evaluation graders and policy models?

\subsection{Setup}
\label{subsec:setup}

\paragraph{Training.}
We use Qwen3-4B-Instruct-2507 and Qwen3-8B (non-thinking)~\citep{qwen3} as base policy models and train them on \texttt{PaperSound-23K} training set.
We report hyperparameters and implementation details in Appendix~\ref{app:sec:hyperparameters}.

\paragraph{Evaluation.}
We evaluate in-domain performance on the \texttt{PaperSound-23K} test set and OOD generalization on PeerReview Bench~\citep{peerreview-bench}, which tests the ability to review Nature-family papers across the physical, biological, and health sciences. We also evaluate on SWE-PRBench~\citep{swe-prbench}, CLAUSE-Legal~\citep{clause}, and StoryFeedback~\citep{storyfeedback}, which test performance on other reviewing tasks: code review, legal contract review, and feedback on story writing, respectively.
We mainly report macro precision, macro recall, and macro F1, except for StoryFeedback, where we report only macro recall (i.e., the error detection rate), following the original work.
For PeerReview Bench, we slightly adapt the evaluation protocol to our problem formulation so that the model reviews only the main paper.
Detailed OOD evaluation is described in Appendix~\ref{app:subsec:ood-benchmarks}.
We use GPT-5.4, a strong frontier model, as an LLM grader to automate these evaluations.
We also conduct an exploratory human preference study comparing PaperCritic with a frontier model at the review and review-point levels.

\paragraph{Baselines.}
We compare against three types of AI reviewers for assessing paper soundness.

\textbf{(1) Models trained with conventional methods.}
For a fair comparison with PaperCritic-4B, we use the same base model and train it using SFT~\citep{cyclereview,deepreview} and RL with different rewards, including a \textit{preference reward}~\cite{reviewrl}, based on holistic preference relative to a human review, and a \textit{similarity reward}~\citep{guide}, based on the ROUGE score against a reference review.
We also train a model using DPO~\citep{goodpoint}, but use soundness and substantiveness pairs consistent with our problem formulation.

\textbf{(2) Larger/reasoning models from the same model family.}
We include Qwen3-32B and Qwen3-235B with or without thinking~\citep{qwen3}.

\textbf{(3) Frontier models.}
We use DeepSeek-V4-Flash-0731~\citep{deepseekv4}, Gemini 3 Flash~\citep{gemini3flash}, Gemini 3.1 Pro~\citep{gemini3pro}, Claude Sonnet 5~\citep{claudesonnet5}, and Claude Opus 5~\citep{claudeopus5}, with high reasoning effort.

\paragraph{Reward models and matcher selection.}
We use a GPT-5.4-distilled Qwen3-4B-Instruct-2507 model as the soundness RM and Qwen3.6-35B-A3B as the substantiveness RM and matcher. Appendix~\ref{app:sec:rm-matcher} provides the detailed selection process.

\subsection{Results}
\label{subsec:results}

We report in-domain results in Table~\ref{tab:main-results} and domain generalization results in Tables~\ref{tab:peerreview-bench} and \ref{tab:cross-task-generalization}.
Our main findings are as follows:

\begin{table}[t]
\centering
\small
\setlength{\tabcolsep}{3pt}
\renewcommand{\arraystretch}{1.08}

\begin{tabularx}{0.95\columnwidth}{l c c c}
\toprule
 & \textbf{Precision} & \textbf{Recall} & \textbf{F1} \\
\midrule

\multicolumn{4}{l}{\textbf{Qwen3 larger/reasoning models}} \\
Qwen3-32B (non-thinking)  & 14.8 & 15.5 & 11.8 \\
Qwen3-235B (non-thinking) & 17.5 & 21.2 & 10.7 \\
Qwen3-32B (thinking)      & 16.5 & 22.6 & 13.5 \\
Qwen3-235B (thinking)     & 19.5 & 34.3 & 19.2 \\
\midrule
\multicolumn{4}{l}{\textbf{Frontier models}} \\
DeepSeek-V4-Flash-0731    & 36.0 & 55.2 & 39.3 \\
Gemini 3 Flash            & 24.0 & 52.1 & 28.7 \\
Gemini 3.1 Pro            & 24.8 & 44.4 & 25.0 \\
Claude Sonnet 5           & 26.0 & 49.4 & 28.9 \\
Claude Opus 5             & 29.3 & \textbf{73.8} & 39.1 \\
\midrule
\multicolumn{4}{l}{\textbf{Trained 4B models}} \\
\textit{Base} \\
Qwen3-4B-Instruct-2507        & 11.3 & 29.6 & 11.2 \\
\rowcolor{LightGrey} \multicolumn{4}{l}{\textit{Models trained with conventional methods}} \\
SFT                           & 23.0 & 25.6 & 16.6 \\
~~+RL w/ preference reward    & 20.6 & 26.5 & 17.1 \\
~~+RL w/ similarity reward    & 24.9 & 34.2 & 23.7 \\
~~+DPO                        & 24.8 & 35.2 & 23.7 \\
\rowcolor{LightGreen} \textit{PaperCritic-4B} &  &  &  \\
~~+RL w/ F1 reward (Ours)     & \textbf{47.4} & 51.4 & \textbf{45.1} \\
\bottomrule
\end{tabularx}
\caption{
\textbf{Performance of AI reviewer models on the \texttt{PaperSound-23K} test set.}
We report macro precision, macro recall, and macro F1 using GPT-5.4 as the LLM grader.
}
\label{tab:main-results}

% \vspace{-5mm}
\end{table}

\begin{table}[t]
\centering
\small
\setlength{\tabcolsep}{4pt}
\renewcommand{\arraystretch}{1.05}

\begin{tabularx}{0.9\columnwidth}{@{}Xccc@{}}
\toprule
\textbf{} & \multicolumn{3}{c}{\textbf{PeerReview Bench}} \\
\cmidrule(l){2-4}
& \textbf{Precision}
& \textbf{Recall}
& \textbf{F1} \\
\midrule
Qwen3-4B-Instruct-2507 & 34.0 & 30.0 & 22.7 \\
SFT                    & 40.9 & 33.3 & 26.8 \\
\rowcolor{LightGreen}
~~+RL (PaperCritic-4B) & \textbf{76.3} & \textbf{49.7} & \textbf{58.7} \\
\bottomrule
\end{tabularx}

\caption{
\textbf{Results on paper reviewing tasks in other domains.}
We evaluate models trained to review ICLR submissions on PeerReview Bench, a benchmark for reviewing Nature-family journal papers.
}
\label{tab:peerreview-bench}
\vspace{-3mm}
\end{table}

\begin{table*}[t]
\vspace{-5mm}
\centering
\small
\setlength{\tabcolsep}{5pt}
\renewcommand{\arraystretch}{1.05}

\begin{tabularx}{0.83\textwidth}{lccccccc}
\toprule
\textbf{} & \multicolumn{3}{c}{\textbf{SWE-PRBench}} & \multicolumn{3}{c}{\textbf{CLAUSE-Legal}} & \textbf{StoryFeedback} \\
\cmidrule(lr){2-4}
\cmidrule(lr){5-7}
\cmidrule(l){8-8}

& \textbf{Precision} & \textbf{Recall} & \textbf{F1}
& \textbf{Precision} & \textbf{Recall} & \textbf{F1}
& \textbf{Recall} \\
\midrule
Qwen3-4B-Instruct-2507 & 7.7 & 8.9 & 7.2 & 13.7 & 15.0 & 13.8 & 78.0 \\
SFT & 3.3 & 1.0 & 1.5 & 9.0 & 11.5 & 8.9 & 74.0 \\
\rowcolor{LightGreen}
~~+RL (PaperCritic-4B)
& \textbf{8.3} & \textbf{17.0} & \textbf{10.5}
& \textbf{14.7} & \textbf{37.3} & \textbf{20.1}
& \textbf{90.0} \\

\bottomrule
\end{tabularx}

\caption{
\textbf{Results on out-of-domain reviewing tasks.}
Models trained on ICLR reviews are evaluated on code review (SWE-PRBench), legal contract review (CLAUSE-Legal), and story feedback (StoryFeedback). Metrics are macro-averaged; StoryFeedback reports only recall. Appendix~\ref{app:subsec:ood-benchmarks} provides evaluation details.
}
\label{tab:cross-task-generalization}
\vspace{-2mm}
\end{table*}

\textbf{PaperCritic-4B outperforms 4B models trained with conventional methods.}
Our model achieves an F1 score of 45.1, which is 21.4 percentage points (pp) higher than that of the best-performing trained 4B baseline. The gain comes from significant improvements in both precision and recall, supporting the effectiveness of our RL training method.

\textbf{PaperCritic-4B outperforms larger/reasoning models from the same model family.}
Our model achieves an F1 score 25.9 pp higher than that of Qwen3-235B in thinking mode. We also observe that, within the Qwen3 model family, larger models and models in thinking mode perform better.

\textbf{PaperCritic-4B outperforms frontier models.}
Our model outperforms DeepSeek-V4-Flash-0731 and Claude Opus 5, the best-performing frontier models, by 5.8 pp and 6.0 pp in F1, respectively.
We observe that frontier models generally have high recall but low precision, implying that their generated reviews are nearly identical to high-quality reference reviews but contain unsupported claims.
In contrast, our model effectively balances recall and precision, which leads to the highest F1 score.
We also report bootstrap confidence intervals for F1 and paired comparisons in Appendix~\ref{app:sec:statistical-uncertainty}.

\textbf{Our training method generalizes effectively to paper reviewing tasks in other domains.}
We evaluate our model on PeerReview Bench and observe a significant improvement over the base model (Table~\ref{tab:peerreview-bench}).
Surprisingly, the F1 score increases by a factor of 2.6, with most of the gain coming from RL training.

\textbf{Our training method also generalizes to out-of-domain reviewing tasks.}
PaperCritic-4B consistently outperforms the base model on SWE-PRBench (code review), CLAUSE-Legal (legal contract review), and StoryFeedback (feedback on story writing) (Table~\ref{tab:cross-task-generalization}).
We observe that the SFT stage has a detrimental effect on performance on these tasks, but RL training compensates for this effect.
In addition, RL training largely improves the coverage (i.e., recall) of reference critiques compared to both the base model and the SFT model.

\begin{figure}[t]
  % \vspace{-8mm}
  \centering
  \includegraphics[width=\columnwidth]{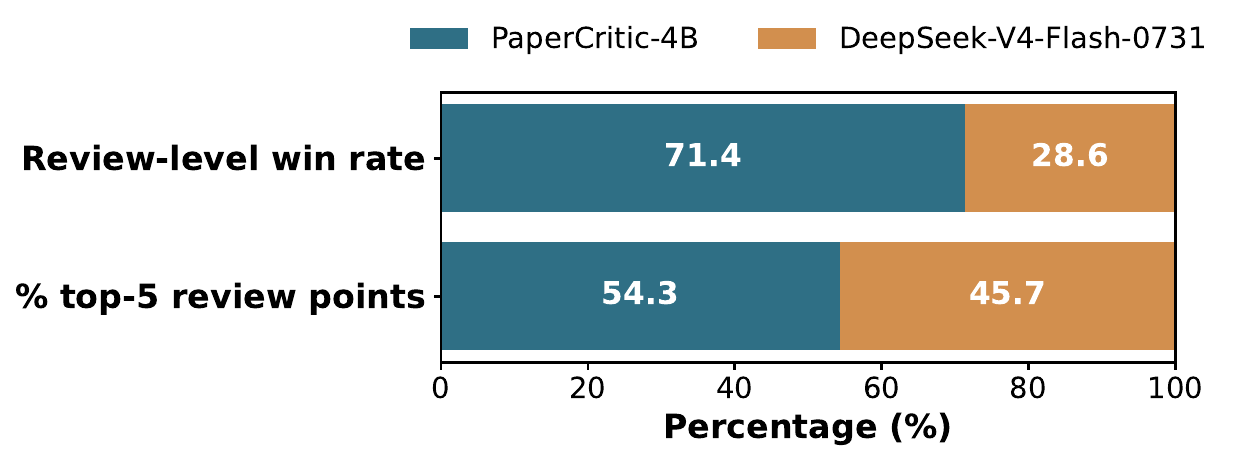}
  \caption{
  \textbf{Human preference for PaperCritic-4B vs. DeepSeek-V4-Flash-0731.}
  We report the review-level win rate and each model's share of the top-5 points selected from both models' reviews.
  }
  \label{fig:human_preference}
  \vspace{-5mm}
\end{figure}

\textbf{Exploratory human preference evaluation.}
Each of seven experts evaluates one paper, comparing anonymized reviews from PaperCritic-4B and DeepSeek-V4-Flash-0731 and selecting the top-5 points from both reviews. The annotation procedure is described in Appendix~\ref{app:subsec:human-preference}.
PaperCritic-4B is preferred in 5/7 comparisons (71.4\%) and contributes 19/35 selected points (54.3\%; Figure~\ref{fig:human_preference}). These results provide preliminary evidence consistent with the automated evaluation.

\subsection{Analysis}
\label{subsec:analysis}

\paragraph{Reward function ablations.}
We verify effectiveness of three components of the reward, soundness, substantiveness, and coverage of review points by ablating each component. We use precision as a reward while ablating the coverage component.
We also add one more ablation, by replacing precision term with one based on reference matching (Table~\ref{tab:ablation}).
The results show that ablating any component leads to consistent drop in F1 score.
To be specific, ablating a matcher or a soundness RM leads to drastic drop in recall and precision, respectively. Ablating a substantiveness RM has a least negative effect, but it still underperforms our original model in recall.
Replacing precision term retains recall score but largely loses precision score.

\begin{table}[t]
\centering
\small
\setlength{\tabcolsep}{5pt}
\renewcommand{\arraystretch}{1.05}

\begin{tabular}{lccc}
\toprule
\textbf{} & \textbf{Precision} & \textbf{Recall} & \textbf{F1} \\
\midrule
PaperCritic-4B             & 47.4 & 51.4 & \textbf{45.1} \\
~~w/o Matcher              & \textbf{65.0} & 10.3 & 13.2 \\
~~w/o Soundness RM         & 13.1 & \textbf{53.0} & 17.6 \\
~~w/o Substantiveness RM   & 52.8 & 35.2 & 36.1 \\
~~w/ Matching-based Precision   & 22.3 & 47.5 & 25.7 \\
\bottomrule
\end{tabular}
\caption{
\textbf{Ablation of our method.}
We ablate the three RL components, two RMs and a matcher, and also replace the precision term with one based on reference matching.
We report the model performance on the \texttt{PaperSound-23K} test set.
}
\label{tab:ablation}
\end{table}

\paragraph{Different LLM grader.}
To mitigate potential model-specific bias in GPT-5.4 as an LLM grader, we also use Gemini 3.8 Flash~\citep{gemini38flash} for evaluation on the \texttt{PaperSound-23K} test set (Table~\ref{tab:gemini-eval}).
The results show a trend consistent with Table~\ref{tab:main-results}: PaperCritic-4B outperforms the other two frontier models in F1.

\begin{table}[t]
\centering
\small
\setlength{\tabcolsep}{5pt}
\renewcommand{\arraystretch}{1.05}

\begin{tabular}{lccc}
\toprule
\textbf{} & \textbf{Precision} & \textbf{Recall} & \textbf{F1} \\
\midrule
DeepSeek-V4-Flash-0731     & 41.5 & 34.1 & 33.3 \\
Claude Opus 5              & 27.3 & \textbf{54.2} & 33.1 \\
\rowcolor{LightGreen} PaperCritic-4B             & \textbf{48.2} & 50.3 & \textbf{44.4} \\
\bottomrule
\end{tabular}
\caption{
\textbf{Performance with a different LLM grader.}
Using Gemini 3.8 Flash as the LLM grader, we evaluate PaperCritic-4B and the two strongest frontier models on the \texttt{PaperSound-23K} test set.
}
\label{tab:gemini-eval}
\vspace{-3mm}
\end{table}

\paragraph{Scaling to larger models.}
To verify whether our training method scales to larger models, we train PaperCritic-8B using Qwen3-8B (non-thinking) as a base model (Table~\ref{tab:8b}).
The results show that both SFT and RL are effective for training the 8B model.
However, PaperCritic-8B underperforms PaperCritic-4B, possibly due to differences in base model quality and insufficient hyperparameter optimization.

\begin{table}[t]
\centering
\small
\setlength{\tabcolsep}{5pt}
\renewcommand{\arraystretch}{1.05}

\begin{tabular}{lccc}
\toprule
\textbf{} & \textbf{Precision} & \textbf{Recall} & \textbf{F1} \\
\midrule
Qwen3-8B         & 13.0 & 4.9 & 4.2 \\
SFT              & 26.0 & 28.7 & 20.0 \\
\rowcolor{LightGreen} ~~+RL (PaperCritic-8B)  & \textbf{38.8} & \textbf{45.1} & \textbf{34.9} \\
\bottomrule
\end{tabular}
\caption{
\textbf{Performance of PaperCritic-8B.}
We train Qwen3-8B (non-thinking) with our method and evaluate PaperCritic-8B on the \texttt{PaperSound-23K} test set.
}
\label{tab:8b}
\vspace{-3mm}
\end{table}

\paragraph{Sensitivity analysis of the soundness RM.}
To examine the effect of soundness RM quality, we use undistilled Qwen3-4B-Instruct-2507 as the lower-quality soundness RM (Table~\ref{tab:rm-matcher}) and train the policy.
Table~\ref{tab:rm-sensitivity} shows that precision is decreased significantly by 27.0 pp.
We also observe that coverage of reference points also decreased by 17.8 pp, which emphasizes the role of using a high-quality soundness RM.

\paragraph{Review style.}
RL induces recurring review styles associated with differences in soundness and reference matching; Appendix~\ref{app:sec:review-style} reports the analysis and examples.

\paragraph{Review point-wise rewards.}
Recursive generation with point-wise rewards underperforms one-shot generation with our review-level F1 reward (37.1 vs.\ 45.1 F1); Appendix~\ref{app:sec:pointwise-rewards} provides the method and results.

\section{Conclusion}
\label{sec:conclusion}

In this work, we introduce PaperCritic, an AI reviewer trained to assess the soundness of papers.
We first formally define paper soundness assessment as jointly maximizing the precision of review points based on soundness and substantiveness and the recall (i.e., coverage) of reference points.
Then, we automatically construct \texttt{PaperSound-23K}, a quality-labeled review point dataset, using LLMs.
We train PaperCritic models on this dataset, including through RL with the F1 score of precision and recall as the reward.
The experimental results show that our model outperforms baselines on the in-domain dataset and effectively generalizes to OOD reviewing tasks.

%%%%%%%
% \newpage
\section*{Limitations}
\paragraph{Scope of paper reviewing.}
We intentionally restrict the scope of AI reviewing to assessing \emph{paper soundness}. 
We do not explicitly optimize for other important purposes of peer review, such as interestingness, clarity, novelty, or the overall accept/reject recommendation, since several of these criteria are inherently more subjective.
Consequently, PaperCritic should be viewed as an assistant for identifying soundness-related concerns rather than a complete replacement for human peer reviewers.
In addition, we exclude review points that rely on external citations and evaluate papers primarily from the information contained in the target paper.
Real-world reviewing often requires comparing a submission against related literature or external domain knowledge, which remains outside the main scope of this work.

\paragraph{Incomplete reference coverage.}
Our recall objective relies on a set of sound and substantive reference review points.
However, neither human-written nor LLM-generated reviews can exhaustively enumerate all valid concerns about a paper.
Therefore, an unmatched generated review point is not necessarily incorrect or unhelpful, and our reference-based recall should be interpreted as coverage of \emph{known} high-quality concerns rather than absolute completeness.

% \section*{Acknowledgments}
% tbd.

\section*{Ethical Considerations}
All human studies complied with our institution's ethics policy, which functions as an Institutional Review Board (IRB)-equivalent review procedure under our institutional arrangements. Every participant gave informed consent before taking part, could stop participating at any point, and received compensation above the local minimum wage for their expertise (see Appendix~\ref{app:sec:human}). PaperCritic is designed to assist human reviewers in identifying soundness concerns; its reviews should not serve as the sole basis for acceptance decisions without human judgment.

\bibliography{custom}

\newpage

\appendix

\section{Human Study/Annotation Details}
\label{app:sec:human}

In this work, we conduct one human study on discovering evaluation criteria of paper soundness assessments (Section~\ref{app:subsec:human-objectives}) and two human annotations on evaluating review point quality (Section~\ref{app:subsec:human-review-quality}) and measuring expert preference on different model-generated reviews (Section~\ref{app:subsec:human-preference}).

All human studies were conducted in compliance with the guidelines established by our institutional ethics committee, which serves a function equivalent to an Institutional Review Board (IRB) under a different local designation. Participants were recruited through snowball sampling via professional and academic networks.
Before participation, all participants were informed of the study procedures, expected task duration (approximately 2--3 hours for each study), and compensation (£69, £46, and £51 for Studies \ref{app:subsec:human-objectives}, \ref{app:subsec:human-review-quality}, and \ref{app:subsec:human-preference}, respectively), and provided informed consent:

\begin{tcolorbox}[
  colback=gray!5,
  colframe=gray!40,
  boxrule=0.4pt,
  arc=1mm,
  left=5pt,
  right=5pt,
  top=4pt,
  bottom=4pt
]
\small
\textbf{Risks and Benefits.} We do not anticipate any specific risks or direct benefits associated with participation in this study. However, if you experience discomfort or an unexpected emotional response during the study, you may stop participating at any time and inform the research team. The research team will respond appropriately and, upon request, may terminate your participation or provide information about additional support resources if needed.

\textbf{Participant Rights.} Your participation in this study is entirely voluntary. If you decide to participate after reading this consent form, you may withdraw your consent or discontinue participation at any time without penalty. You may also choose not to participate or decline to answer any question. The results of this study may be presented at scientific or professional conferences or published in academic journals. Any presentations or written materials resulting from this study will protect your personal information and will not identify you individually.

\textbf{Privacy and Data Protection.} Data collected through this study will be used for research purposes. Where possible, personally identifiable information will be removed or anonymized. The data will be stored securely in an access-controlled environment and will be accessible only to authorized members of the research team. After the study is completed, personal information will continue to be handled securely, and data that are no longer necessary will be deleted or otherwise appropriately disposed of.

\textbf{Contact Information.} If you have any questions, concerns, or complaints regarding the study procedures, potential risks or benefits, or your rights as a participant, please contact the principal investigator (contact).

\vspace{0.5em}

If you agree to participate in this study, please select ``I agree.'' If you do not agree, you may decline participation by informing the researcher.
\end{tcolorbox}

\subsection{Discovering Evaluation Criteria of Paper Soundness Assessments}
\label{app:subsec:human-objectives}

\paragraph{Human study process.}
We recruit two research scientists and five PhD students, where the qualification criterion is more than 3 years of AI research in their fields, which span on reinforcement learning, natural language processing, machine learning theory, computer vision, and speech processing.
Once we recruit human experts, we randomly sampled 100 submissions in ICLR 2020--2023 for each human expert based on their relevant fields and let them select the most relevant work.
Then, human experts spend $\sim$2 hours for evaluating overall quality of given review points on a 5-point scale and justifying their evaluations. Based on this annotation, we also let them to write evaluation criteria of paper soundness assessments.
Once we collect the annotations, we find that we additionally need to clarify the exact meaning of human experts' justifications, so we conduct $\sim$1 hour of interviewing for every human expert and analyze their responses.

\paragraph{Results.}
We first filtered out human experts' justifications where they assume paper reviewing not as assessing soundness of papers, which includes assessing interestingness, novelty, or clarity of papers or providing constructive feedback.
Then, we observe two universal and complete criteria that all human experts have:

\textbf{(1) Soundness.}
They give low quality score if a review point contains an incorrect premise or its claim is not sufficiently supported by evidence (i.e., a review point is \textit{unsound}). During interviewing, we also confirm that high-scoring review points are all \textit{sound}.

\textbf{(2) Substantiveness.}
They give high quality score if a review point raises a critical concern on the paper's main claims (i.e., a review point is \textit{substantive}).

We finally confirm that all high-scoring review points are sound and substantive and no low-scoring review points are sound and substantive.
As a result, we use these two evaluation criteria as an objective of assessing soundness of papers.

\subsection{Evaluating Review Point Quality}
\label{app:subsec:human-review-quality}

\paragraph{Human annotation process.}
We recruit three PhD students who did more than 3 years of AI research in their fields.
Once we recruit human annotators, we give them randomly sampled 100 submissions in ICLR 2020--2023 and they select total seven submissions.
Then, they annotate soundness and substantiveness of human-written and LLM-generated review points of the selected seven submissions. Specifically, we follow the same scoring scale described in Section~\ref{subsec:dataset-construction}.
Since our goal includes measuring human--human agreement, all three humans annotate the same review points.

\subsection{Measuring Preference of Different AI Reviewers}
\label{app:subsec:human-preference}

\paragraph{Human annotation process.}
We recruit one postdoctoral researcher and six PhD students, each with more than 3 years of AI research experience. From a randomly sampled pool of 100 ICLR 2020–2023 submissions, each expert selects the paper most relevant to their expertise, yielding seven expert-selected papers with one annotator per paper.
For each paper, we generate reviews using DeepSeek-V4-Flash-0731 and PaperCritic-4B. The expert ranks the two anonymized reviews by overall preference and selects the top-5 review points from their combined pool, regardless of which model generated them.

\section{\texttt{PaperSound-23K} Details}
\label{app:sec:add-stats}

\paragraph{Licensing.}
The papers and reviews included in this work are all obtained from OpenReview, under a CC-BY 4.0 license, and our use of them is consistent with the OpenReview terms of use: https://openreview.net/legal/terms.

\paragraph{Paper Statistics.}
The complete dataset contains 989 ICLR papers from 2020--2023.
The year distribution is 167 (16.89\%) in 2020, 192 (19.41\%) in 2021, 233 (23.56\%) in 2022, 397 (40.14\%) in 2023.
Of all papers, 392 (39.64\%) were accepted and 597 (60.36\%) were rejected.
The recorded research-area metadata contain 242 distinct labels. The most frequent named labels are reinforcement learning (84; 8.49\%), robustness/safety (50; 5.06\%), graph/structure (46; 4.65\%), vision (40; 4.04\%). The ten largest named labels together account for 358 papers (36.20\%); 121 papers (12.23\%) have the recorded label ``Other,'' and the remaining 510 papers (51.57\%) are distributed across 231 additional labels. Table~\ref{tab:paper-statistics} summarizes these distributions.

\begin{table}[h]
\centering
\small
\resizebox{\columnwidth}{!}{%
\begin{tabular}{@{}llrr@{}}
\toprule
\textbf{Dimension} & \textbf{Category} & \textbf{Papers} & \textbf{Percentage (\%)} \\
\midrule
Year & 2020 & 167 & 16.89 \\
 & 2021 & 192 & 19.41 \\
 & 2022 & 233 & 23.56 \\
 & 2023 & 397 & 40.14 \\
\midrule
Acceptance & Accepted & 392 & 39.64 \\
 & Rejected & 597 & 60.36 \\
\midrule
Research area & RL & 84 & 8.49 \\
 & Robustness/Safety & 50 & 5.06 \\
 & Graph/Structure & 46 & 4.65 \\
 & Vision & 40 & 4.04 \\
 & NLP/LLM & 28 & 2.83 \\
 & Generative & 26 & 2.63 \\
 & Theory & 25 & 2.53 \\
 & Deep Learning & 23 & 2.33 \\
 & Federated Learning & 20 & 2.02 \\
 & Optimization & 16 & 1.62 \\
 & Other (recorded label) & 121 & 12.23 \\
 & Remaining labels (231) & 510 & 51.57 \\
\midrule
Total & All papers & 989 & 100.00 \\
\bottomrule
\end{tabular}
}
\caption{
\textbf{Paper-level statistics for the full dataset.}
The ten largest named research-area labels are shown individually. The recorded \texttt{Other} label is separate from the pooled remaining labels.
}
\label{tab:paper-statistics}
\end{table}

\begin{figure*}[t]
    \centering

    \begin{subfigure}[t]{0.33\textwidth}
        \centering
        \includegraphics[
            width=\linewidth
        ]{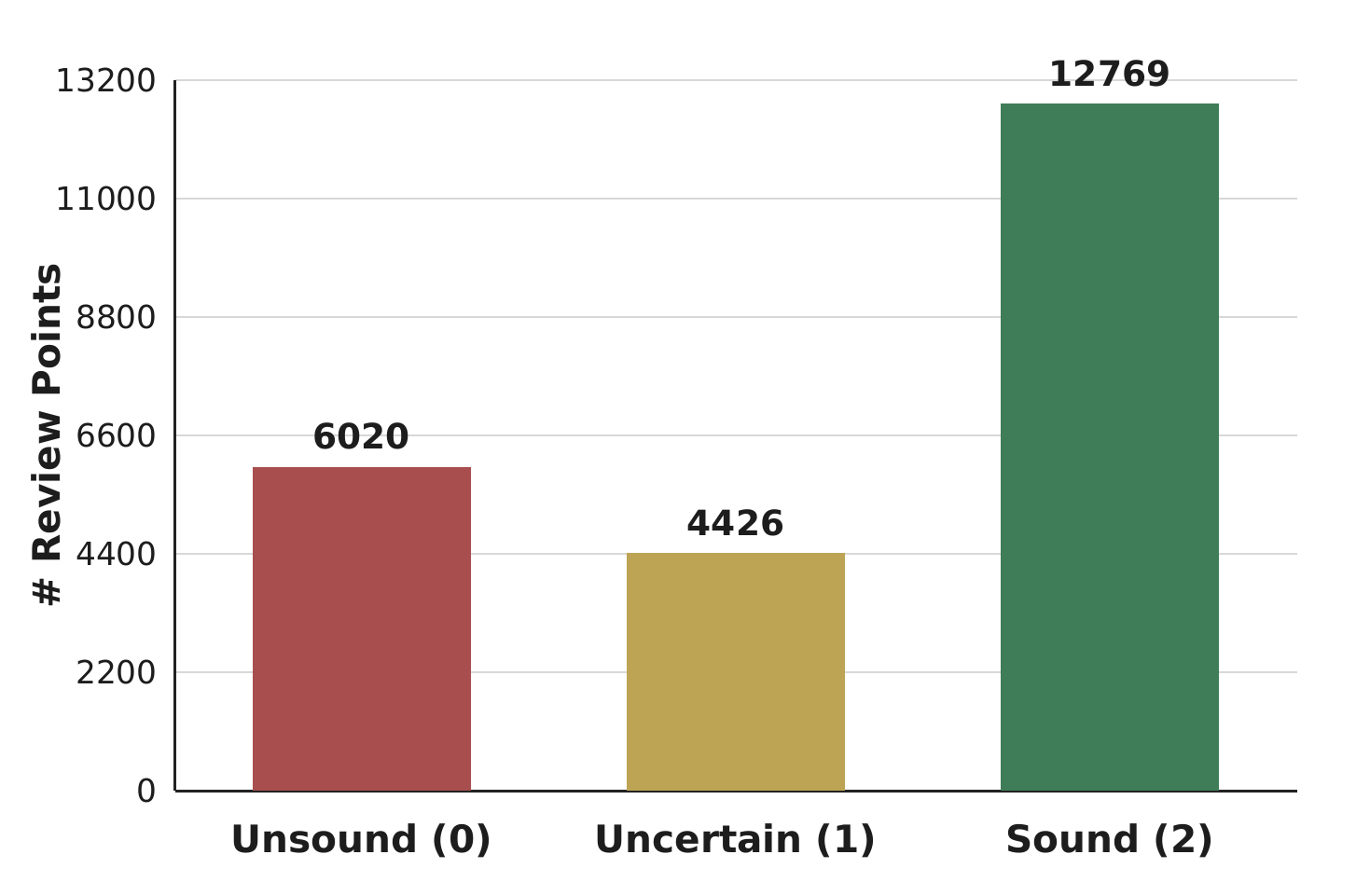}
        \caption{}
        \label{subfig:soundness_distribution}
    \end{subfigure}
    \hfill
    \begin{subfigure}[t]{0.27\textwidth}
        \centering
        \includegraphics[
            width=\linewidth
        ]{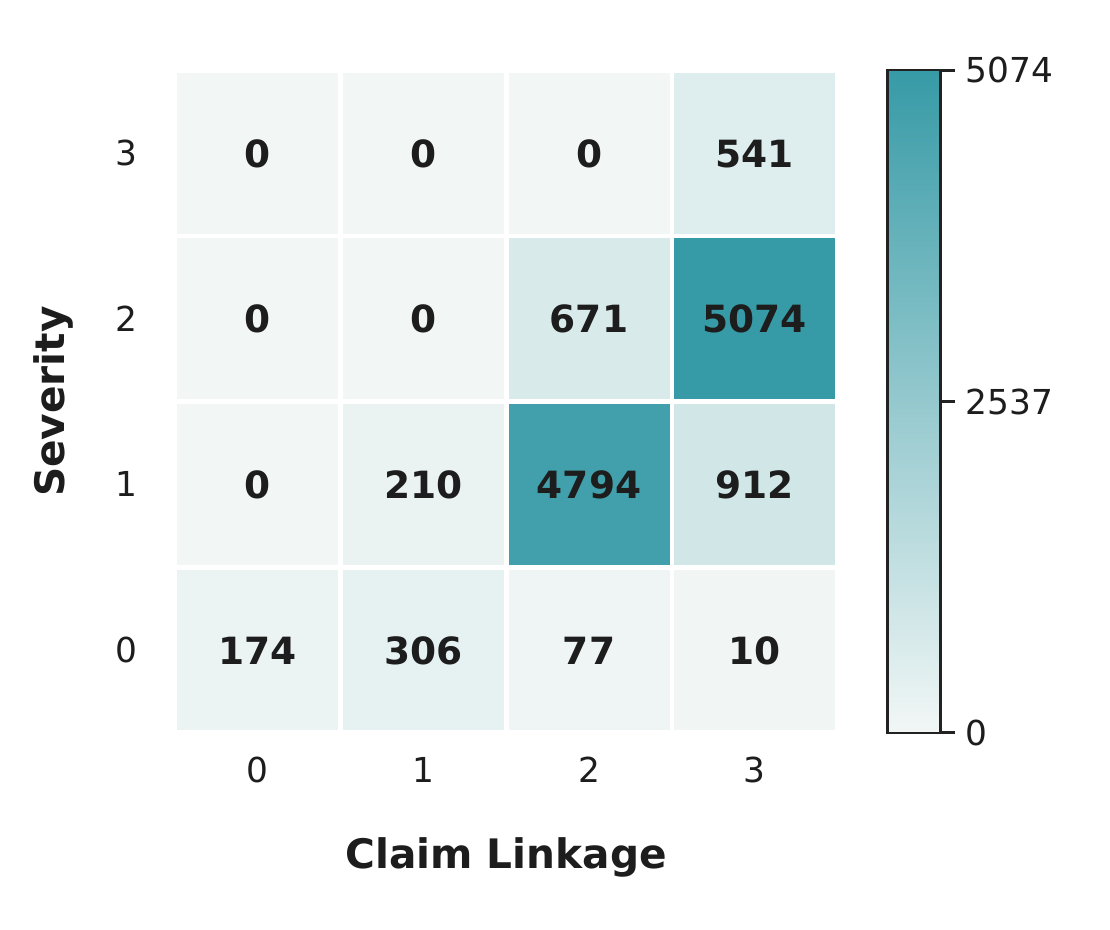}
        \caption{}
        \label{subfig:linkage_severity_heatmap}
    \end{subfigure}
    \hfill
    \begin{subfigure}[t]{0.38\textwidth}
        \centering
        % \vspace{1mm}
        \includegraphics[
            width=\linewidth
        ]{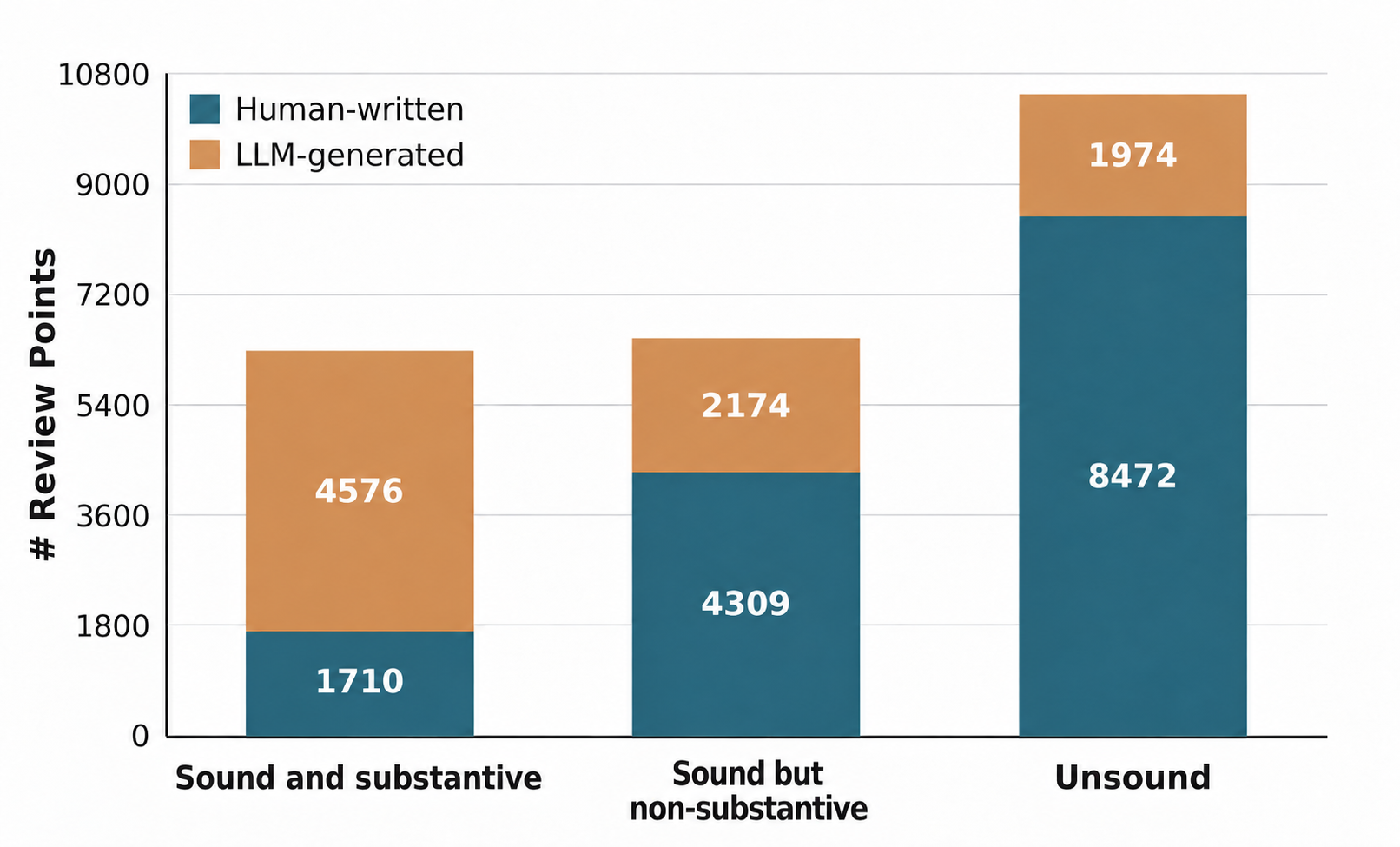}
        \caption{}
        \label{subfig:quality_distribution}
    \end{subfigure}
    \caption{
    \textbf{Quality score and label distribution of \texttt{PaperSound-23K}.}
    (a) Soundness score (3-point scale) distribution.
    (b) Claim linkage and severity score (4-point scale) distribution of sound review points.
    (c) Comparison of quality label distribution of human-written and LLM-generated review points.
    }
    \label{fig:scored_review_point_statistics}
\end{figure*}

\paragraph{Quality score and label distribution.}
We report the quality score distribution in Figure~\ref{subfig:soundness_distribution} and \ref{subfig:linkage_severity_heatmap}.
55.0\% of the review points are sound, 25.9\% are clearly unsound, and 19.1\% are uncertain to judge.
Regarding substantiveness, 94.6\% of the sound review points have $\mathrm{ClaimLinkage}\geq 2$, but only 49.2\% of the sound review points have $\mathrm{Severity}\geq 2$.
We also compare the quality label distribution of human-written and LLM-generated review points in Figure~\ref{subfig:quality_distribution}. 72.8\% of sound and substantive review points are LLM-generated, and 81.1\% of unsound review points are human-written.

\section{Hyperparameters and Implementation Details}
\label{app:sec:hyperparameters}

In this section, we describe training details of the experiments, which are categorized into three types: SFT, DPO, and RL.
RL includes three types of rewards: preference reward, similarity reward, and our F1 reward.

SFT trains on reference reviews containing up to nine high-quality review points, with loss computed only on the assistant response.
Both DPO and RL initialize from the SFT checkpoint.
All methods use full-parameter fine-tuning with BF16, FSDP, activation checkpointing, AdamW, zero weight decay, gradient clipping at 1.0, and random seed 42.
SFT uses a maximum total sequence length of 16,384 tokens. DPO and RL instead allow 16,384 prompt tokens plus 2,304 completion tokens.
We summarize main hyperparameters in Table~\ref{tab:training-hyperparameters}.

\begin{table}[h]
\centering
\small
\setlength{\tabcolsep}{3pt}
\renewcommand{\arraystretch}{1.05}
\begin{tabularx}{\columnwidth}{@{}>{\hsize=1.28\hsize\linewidth=\hsize\raggedright\arraybackslash}X *{3}{>{\hsize=.9066667\hsize\linewidth=\hsize\centering\arraybackslash}X}@{}}
\toprule
\textbf{Hyperparameter} & \textbf{SFT} & \textbf{DPO} & \textbf{RL} \\
\midrule
Learning rate
& $1.3\times10^{-5}$
& $5\times10^{-6}$
& $5\times10^{-6}$ \\

Paper epochs
& 3
& 2
& 2 \\

Effective batch size
& 16 reviews
& 16 pairs
& 30 reviews \\

Warmup ratio
& 0.05
& 0.03
& 0.03 \\

LR schedule
& Constant after warmup
& Linear decay
& Linear decay \\
\bottomrule
\end{tabularx}
\caption{
\textbf{Training hyperparameters for SFT, DPO, and RL.}
We use the same hyperparameters for RL with three different rewards.
}
\label{tab:training-hyperparameters}
\end{table}

\subsection{RL}
\label{app:subsec:rl}

We use GRPO with the DAPO token-normalized loss, batch-level reward standard-deviation scaling, and a KL coefficient of $0.003$.
Each update samples six reviews for each of five papers.
Rollouts use temperature $0.7$, top-$p=0.8$, and top-$k=20$.
The implementation allocates five NVIDIA A100-80GB GPUs to the policy and three to reward evaluation.

\paragraph{Preference baseline.}
Qwen3.6-35B-A3B without thinking, at temperature zero, compares the generated review against an equally sized sample of human-written points from the same paper, without high-quality filtering. Sampling is without replacement unless the human pool is too small. The judge receives the paper, extracted claims/evidence, paper year, and anonymized reviews A/B in seeded randomized order. The prompt asks which review is better based on soundness and substantiveness. It penalizes false premises, repetition, and generic criticism, discourages preference for longer or more polished prose, and permits ties when neither review has a clear advantage. A generated-review win, tie, or loss yields reward $1$, $0.5$, or $0$, respectively.

\paragraph{Similarity baseline.}
The reward is summary-level ROUGE-L F-measure, implemented as \texttt{rougeLsum} in \texttt{rouge-score} 0.1.2 with stemming enabled. Points are whitespace-normalized and joined by newlines. References contain all human- and model-written high-quality (i.e. sound and substantive) points.

\subsection{DPO}
\label{app:subsec:dpo}

We use sigmoid DPO with $\beta=0.1$ and a frozen SFT reference model.
The training set contains 5,238 preference pairs from 873 papers, equally divided between soundness and substantiveness pairs.
Each pair has equal-length reviews; the preferred review improves the targeted criterion by two points while preserving the count satisfying the other criterion.
For fair comparison with RL, we use three pairs per paper per epoch which matches to six rollouts in RL.
We use four NVIDIA A100-80GB GPUs for implementation.

\section{Baselines and Evaluation Details}
\label{app:sec:baseline-eval-details}

\subsection{Inputs, Inference, and Output Lengths}
\label{app:subsec:eval}

\paragraph{Instructions and input content.}
All in-domain models review the same papers using the same core instruction: produce a numbered list of sound, substantive criticisms that could undermine a main claim, excluding novelty and minor presentation concerns. Inputs contain the paper's title, abstract, and section text, without reference reviews.
However, the prompts and available context are not strictly identical. Trained SFT/DPO/RL models use the original SFT instruction and a 14,336-token input cap, retaining the paper prefix after reserving instruction/template space; four papers are truncated.
Unadapted base, larger-Qwen, and frontier models receive the full rendered input, with additional instructions to produce at most eight points and target 100 words per point, and never exceed 132 words per point. This selection is made for fair comparison with the trained models; 100 words are an average number of words of PaperCritic-4B's retained review points, and 132 words are their 97.5th-percentile length.

\paragraph{Decoding and parsing.}
Generation is one-shot. Local settings are temperature $0.7$, top-$p=0.8$, top-$k=20$, repetition penalty $1.0$, and base seed 42. API runs request the same temperature, top-$p$, top-$k$, and seed where supported.
Thinking is disabled for trained and base models; larger Qwen models are tested in both modes, and frontier models request high reasoning effort. Completion caps are listed in Table~\ref{tab:baseline-output-lengths}; API budgets may include hidden reasoning.

The parser removes reasoning wrappers and code fences, splits numbered/bulleted items with their continuation lines, and retains only the first eight points. The trained-model parser additionally permits paragraph/line fallbacks and retains duplicates. The length-controlled suite requires list formatting and validates retained-point length, exact duplicates, and leaked reasoning, allowing up to two standard repair retries. It does not truncate individual points or select responses by review quality.

\begin{table*}[t]
\centering
\small
\setlength{\tabcolsep}{4pt}
\begin{tabular}{lrrrrr}
\toprule
& Completion & \multicolumn{2}{c}{Points/paper} & \multicolumn{2}{c}{Visible tokens/paper} \\
\cmidrule(lr){3-4}\cmidrule(lr){5-6}
Model & cap & Parsed & Retained & Full response & Retained points \\
\midrule
Qwen3-4B-Instruct-2507 & 16,384 & 8.00 & 8.00 & 852.2 & 836.2 \\
SFT & 2,048 & 6.58 & 5.80 & 620.0 & 544.1 \\
~~+DPO & 2,048 & 6.02 & 6.02 & 538.8 & 526.7 \\
~~+RL w/ preference reward & 2,048 & 5.02 & 5.02 & 506.0 & 495.9 \\
~~+RL w/ similarity reward & 2,048 & 8.74 & 7.96 & 566.6 & 505.6 \\
~~+RL w/ F1 reward (Ours) & 2,048 & 8.86 & 7.98 & 1,392.3 & 1,242.4 \\
\midrule
Qwen3-32B (non-thinking) & 16,384 & 8.00 & 8.00 & 507.0 & 491.0 \\
Qwen3-32B (thinking) & 16,384 & 8.00 & 8.00 & 604.3 & 582.9 \\
Qwen3-235B-A22B (non-thinking) & 8,192 & 8.00 & 8.00 & 530.8 & 514.8 \\
Qwen3-235B-A22B (thinking) & 8,192 & 8.00 & 8.00 & 524.5 & 502.7 \\
\midrule
DeepSeek-V4-Flash-0731 & 16,384 & 7.54 & 7.42 & 918.4 & 872.7 \\
Gemini 3 Flash & 16,384 & 6.36 & 6.36 & 870.5 & 857.8 \\
Gemini 3.1 Pro & 32,768 & 5.74 & 5.74 & 741.3 & 730.0 \\
Claude Sonnet 5 & 16,384 & 7.94 & 7.94 & 761.0 & 745.2 \\
Claude Opus 5 & 16,384 & 8.00 & 8.00 & 1,273.2 & 1,257.5 \\
\bottomrule
\end{tabular}
\caption{
\textbf{Mean output sizes on the \texttt{PaperSound-23K} test set.}
Visible text is re-tokenized with the Qwen3-4B-Instruct-2507 tokenizer, without special tokens or hidden reasoning; these are text-length measurements, not billed-token counts. Full-response tokens include list formatting and discarded points; retained tokens sum the evaluated point texts.
}
\label{tab:baseline-output-lengths}
\end{table*}

\subsection{OOD Evaluation}
\label{app:subsec:ood-benchmarks}

\paragraph{Shared inference settings.}
Base Qwen3-4B-Instruct-2507, SFT, and PaperCritic receive identical inputs and domain instructions within each benchmark. They generate numbered lists without a per-point word limit; only the first eight points are evaluated. Thinking is disabled, top-$k=20$, repetition penalty is $1.0$, and the seed is 42 plus the instance index. PeerReview Bench and SWE-PRBench use temperature $0.7$, top-$p=0.8$, and a 16,384-token input cap; CLAUSE-Legal uses $0.6$, $0.9$, and 24,576; StoryFeedback uses $0.6$, $0.9$, and 8,192.
Inputs are prefix-truncated when necessary, with legal authority text reserved in full. Completion caps increase from 2,048 to 4,096 and then 8,192 only after a length cutoff. Remaining cutoff outputs and duplicate points are retained.

\paragraph{PeerReview Bench.}
We evaluate the fixed first 50 papers (IDs 1--50), with 473 reference review items.
Models receive main-paper text and available figure captions, without image pixels or supplementary files, and the same paper soundness instruction as in-domain generation.
GPT-5.4 grades precision as the fraction of points rated \emph{Correct}, \emph{Significant}, and supported by \emph{Sufficient} evidence. Reference coverage requires a generated point with the same subject and argument, allowing different evidence.
References requiring appendix information are not excluded: all 473 items remain, despite main-paper-only model inputs.

\paragraph{SWE-PRBench.}
We use the \texttt{eval\_split} configuration of \texttt{foundry-ai/swe-prbench}, with 598 human comments.
Models receive PR metadata and the diff, without repository browsing. The prompt requests concrete correctness, reliability, security, performance, or integration issues grounded in the supplied change, not style-only criticism.
The saved judge is GPT-5.4, receiving generated/human comments and up to 12,000 characters of selected diff. Precision is the fraction of generated comments classified as \texttt{CONFIRMED} against a human issue; plausible but reference-unmatched issues do not count. Recall measures human comments classified as \texttt{CAUGHT}.

\paragraph{CLAUSE-Legal.}
We use a seed-42 sample of 100 positively annotated perturbed contracts: 50 CUAD and 50 ContractNLI.
The reference sets contain 298 legal perturbations. The prompt requests contract--law conflicts and additionally supplies each relevant \texttt{law\_citation} and \texttt{scraped\_snippet\_1}; perturbation explanations are withheld from generation.
One GPT-5.4 call per nonempty review performs many-to-many matching using generated points and reference perturbation descriptions, including available explanations, justifications, and law evidence. Precision is the fraction of generated points matched to at least one reference; recall counts uniquely covered references.

\paragraph{StoryFeedback.}
We use 100 corrupted inputs sampled with seed 42 from \texttt{igen/data/generated\_feedback.jsonl}, spanning 92 original stories; uncorrupted inputs are excluded.
The prompt requests grounded, specific, actionable feedback on story coherence, clarity, fluency, style, or needed additions/deletions, permitting \emph{The text is perfect as-is.}
GPT-5.4 evaluates the entire feedback list. There is no point-level reference set or reported precision: the reported recall score is binary error detection, indicating whether any feedback addresses the biggest problem. This judge sees only the corrupted story and feedback.

\section{Reward Models and Matcher Selection Details}
\label{app:sec:rm-matcher}

We first compare the ability of Qwen3-14B~\citep{qwen3} and Qwen3.6-35B-A3B~\citep{qwen36} as RMs and a matcher. To do that, we reserve 50 papers and their corresponding human- and LLM-written review points from a \texttt{PaperSound-23K} training set and use those as held-out split.
We observe that Qwen3.6-35B-A3B has stronger correlation with GPT-5.4 (Table~\ref{tab:rm-matcher}).
However, both models show weak correlation on soundness evaluation, so we distill soundness judgments of GPT-5.4 to Qwen3-4B-Instruct-2507.

\paragraph{Distilled soundness RM.}
We fine-tune Qwen3-4B-Instruct-2507 to reproduce GPT-5.4 soundness judgments and rationales. From the 839 papers in a \texttt{PaperSound-23K} training set (excluding 50 held-out papers), we construct a 16,600-example training mixture: 45\% existing human- and LLM-written review points, 15\% SFT-policy-generated points, and 40\% DPO-policy-generated points.
Sampling includes oversampling where necessary and targets a 40:20:40 distribution over unsound, uncertain, and sound labels. After excluding examples exceeding the 16,384-token sequence limit, 16,288 training examples remain. We perform full-parameter SFT for three epochs with learning rate $8\times10^{-6}$ and effective batch size 16, applying token-level cross-entropy only to the assistant's JSON response containing the soundness score, premise reconstruction, and factuality rationale.

\paragraph{Results.}
We verify that this distilled model has significantly stronger correlation than the undistilled model (Table~\ref{tab:rm-matcher}).
As a result, we use distilled Qwen3-4B-Instruct-2507 as a soundness RM and Qwen3.6-35B-A3B as a substantiveness RM and a matcher.

\begin{table}[h]
\centering
\small
\setlength{\tabcolsep}{3pt}
\renewcommand{\arraystretch}{1.05}

\begin{tabularx}{\columnwidth}{@{}Yccc@{}}
\toprule
\textbf{Candidate models} & \makecell{\textbf{Sound-}\\\textbf{ness}} & \makecell{\textbf{Substan-}\\\textbf{tiveness}} & \makecell{\textbf{Ref.-}\\\textbf{Matching}} \\
\midrule
Qwen3-4B-\allowbreak Instruct-2507 & -0.059 & --- & --- \\
Qwen3-14B              & 0.107 & 0.574 & 0.429 \\
Qwen3.6-35B-A3B        & 0.244 & \textbf{0.596} & \textbf{0.613} \\
Distilled Qwen3-4B-\allowbreak Instruct-2507 & \textbf{0.407} & N/A & N/A \\
\bottomrule
\end{tabularx}
\caption{\textbf{Performance as reward models and a matcher.}
We report Pearson's correlation of candidate models and GPT-5.4 on evaluating soundness, substantiveness, and reference-matching.
We report substantiveness correlation as an average of claim linkage and severity correlations and reference-matching correlation as an average of relatedness and specificity correlations.
}
\label{tab:rm-matcher}
\end{table}

\section{Statistical Uncertainty}
\label{app:sec:statistical-uncertainty}

We report single-run results and quantify test-paper sampling uncertainty for PaperCritic-4B and five frontier baselines on the same \texttt{PaperSound-23K} test papers (Table~\ref{tab:bootstrap}).
Using the saved GPT-5.4 evaluations, we construct 95\% percentile bootstrap confidence intervals from 100,000 resamples (seed 42), each containing 100 papers sampled with replacement. The same sampled paper indices are used across models, preserving pairing when estimating F1 differences. Reported point estimates are computed on the original test set, not averaged or maximized over bootstrap replicates.

\begin{table*}[t]
\centering
\small
\setlength{\tabcolsep}{5pt}
\begin{tabular}{lrrrr}
\toprule
\textbf{Model} & \textbf{F1 (\%)} & \textbf{95\% CI} & \textbf{$\Delta$ F1 (pp)} & \textbf{Paired 95\% CI} \\
\midrule
PaperCritic-4B & 45.14 & [39.20, 50.81] & --- & --- \\
DeepSeek-V4-Flash-0731 & 39.32 & [33.56, 44.82] & $+5.82$ & [$-0.75$, $+12.62$] \\
Claude Opus 5 & 39.10 & [34.10, 44.07] & $+6.03$ & [$+0.32$, $+11.82$] \\
Gemini 3 Flash & 28.67 & [24.09, 33.12] & $+16.47$ & [$+10.38$, $+22.38$] \\
Gemini 3.1 Pro & 25.04 & [19.38, 30.74] & $+20.10$ & [$+14.68$, $+25.66$] \\
Claude Sonnet 5 & 28.86 & [23.76, 33.93] & $+16.27$ & [$+10.26$, $+22.01$] \\
\bottomrule
\end{tabular}
\caption{
\textbf{Macro F1 and bootstrap confidence intervals on the \texttt{PaperSound-23K} test set.}
Differences are PaperCritic-4B minus each baseline, in percentage points (pp).
Intervals are pointwise and not adjusted for multiple comparisons.
}
\label{tab:bootstrap}
\end{table*}

The results show that PaperCritic-4B's paired improvement intervals exclude zero against Claude Opus 5, Gemini 3 Flash, Gemini 3.1 Pro, and Claude Sonnet 5, which further supports the effectiveness of our model.

\section{Analysis of Review Style}
\label{app:sec:review-style}

We analyze whether training induces specific review styles and whether these are associated with review quality (Table~\ref{tab:style-narrowing}).
Through manual inspection, we identify three recurring patterns: points containing ``can you'', points containing ``but it is unclear that'', and points identifying inconsistencies in notation or equations.
The first two patterns are associated with higher soundness and reference-matching rates, whereas the last is associated with lower rates on both criteria.
We leave investigation of the causes of style narrowing and mitigation strategies to future work.

\begin{table}[t]
\centering
\small
\setlength{\tabcolsep}{3.5pt}
\renewcommand{\arraystretch}{1.08}

\begin{tabularx}{\columnwidth}{@{}Xcc|cc@{}}
\toprule
& \multicolumn{2}{c}{\textbf{Freq.}}
& \multicolumn{2}{c}{\textbf{Quality Assoc.}} \\
\cmidrule(lr){2-3}
\cmidrule(l){4-5}
\textbf{Pattern}
& \textbf{SFT}
& \textbf{RL}
& \textbf{Sound}
& \textbf{Ref.-matched} \\
\midrule

``Can you''
& 0.0 & \textbf{71.9}
& +15.5 & +6.0 \\

``But it is unclear that''
& 0.3 & \textbf{41.9}
& +10.8 & +7.1 \\

Notation inconsistency
& 0.3 & \textbf{9.5}
& --25.1 & --12.5 \\

\bottomrule
\end{tabularx}
\caption{
\textbf{Style narrowing after RL and its association with review point quality.}
Frequency is the percentage of review points containing each pattern.
Quality columns report the percentage-point difference between RL review points \emph{with} versus \emph{without} the pattern.
}
\label{tab:style-narrowing}
\vspace{-3mm}
\end{table}

The following excerpts illustrate the three patterns; omissions are marked
with $[\ldots]$.

\paragraph{``Can you'': requesting concrete evidence.}
\emph{Do We Really Need Labels for Backdoor Defense?}
\begin{quote}
\small
can you provide quantitative evidence $[\ldots]$ that these contrastive
adversaries actually lie in the backdoor region/cluster and not just in some
general adversarial direction
\end{quote}
The point requests missing feature-space validation of the proposed mechanism,
rather than asserting that the mechanism is false. It was judged sound and
matched to a reference concern.

\paragraph{``But it is unclear that'': questioning an inference.}
\emph{Domain-Wise Adversarial Training for Out-of-Distribution Generalization.}
\begin{quote}
\small
The ``instance reweighting'' argument in Proposition 1 (Eq. 11) is stated as
an existence claim but it is unclear that it actually implies a practical
connection that makes DAT a better proxy for IRMv1/IRM in deep networks.
\end{quote}
The point distinguishes a formal relationship from evidence of its practical
significance. Its request for validation was judged sound and matched to
references questioning the DAT--IRM connection.

\paragraph{Notation inconsistency: an unsupported correction.}
\emph{$\mathrm{SO}(2)$-Equivariant Reinforcement Learning.}
\begin{quote}
\small
Several core definitions/notations appear inconsistent in ways that affect
the stated equivariance constraints $[\ldots]$ Definition 4.1's
transition-invariance condition is written as $T(s,a,s')=T(gs,ga,gs')$ but it
should be $T(s,a,s')=T(gs,ga,gs')$ under the stated group action $[\ldots]$
\end{quote}
The proposed correction is identical to the original expression, so this
example does not establish the alleged inconsistency. The point was judged
unsound and did not match any reference.

\section{Results of Sensitivity Analysis of Soundness RMs}
\label{app:sec:sensitivity}
As we describe in Section~\ref{subsec:analysis}, the result is provided in Table~\ref{tab:rm-sensitivity}.

\begin{table}[h]
\centering
\small
\setlength{\tabcolsep}{3pt}
\renewcommand{\arraystretch}{1.05}

\begin{tabularx}{\columnwidth}{@{}Yccc@{}}
\toprule
\textbf{Soundness RM} & \textbf{Precision} & \textbf{Recall} & \textbf{F1} \\
\midrule
Current RM                      & \textbf{47.4} & \textbf{51.4} & \textbf{45.1} \\
Qwen3-4B-Instruct-2507          & 20.4 & 33.6 & 19.7 \\
\bottomrule
\end{tabularx}
\caption{
\textbf{Sensitivity analysis of soundness RMs.}
We use undistilled Qwen3-4B-Instruct-2507 as a soundness RM and compare the training result with the one that uses the GPT-5.4-distilled RM on the \texttt{PaperSound-23K} test set.
}
\label{tab:rm-sensitivity}
\end{table}

\section{Details of Training Recursive Review Points Generation Model}
\label{app:sec:pointwise-rewards}

To provide finer-grained feedback than a single review-level reward, we adapt the one-shot formulation to recursive generation and assign rewards to individual review points.
Given a paper $x$ and previously generated points $h_t = (p_1, \ldots, p_{t-1})$,
the policy generates either one additional review point or a termination signal:
\[
\begin{gathered}
a_t \sim \pi_\theta(\cdot \mid x, h_t),\\
a_t \in \{\text{review point}, \texttt{<STOP>}\}.
\end{gathered}
\]
The prefix is initially empty.
Each generated point is appended to the prefix, and the model is instructed to identify a sound, substantive weakness not already covered.
Generation terminates upon producing \texttt{<STOP>} or reaching a predefined generation limit.
Reference reviews are used for training supervision and reward computation, but are not provided as part of the policy input.

\subsection{Supervised Fine-Tuning}

We adapt the same point selection and claim-diversity-aware ordering used for one-shot SFT to construct next-point prediction examples.
For an ordered sequence $(p_1,\ldots,p_m)$, we construct examples
\[
\begin{aligned}
(x,\varnothing) &\rightarrow p_1,\\
(x,p_1) &\rightarrow p_2,\\
&\phantom{\rightarrow}\vdots\\
(x,p_1,\ldots,p_{m-1}) &\rightarrow p_m.
\end{aligned}
\]
We additionally construct a \texttt{<STOP>} target after the complete target sequence has been supplied.
This terminal example is omitted when the sequence has been truncated by the maximum-point limit, so that truncation is not incorrectly treated as exhaustion of useful feedback.
Training minimizes the negative log-likelihood of the target point or stop signal, with the paper and preceding points serving only as conditioning context.
The claim-diverse ordering is resampled across training epochs.

\subsection{Recursive Reinforcement Learning}

Starting from the recursively trained SFT model, we alternate between \textbf{state-buffer construction} and \textbf{GRPO optimization}.

For buffer construction, we take a batch of training papers and sample one recursive trajectory per paper using the current policy, holding its parameters fixed during this phase.
At each step, we store the paper, its generated prefix, and the reference-coverage information associated with that prefix.
Nonterminal points are appended without filtering them by reward, allowing subsequent states to include imperfect model-generated feedback.
A matcher identifies reference points covered by each sampled point and updates the remaining-reference set for subsequent states.
Trajectories continue until the stop signal or the generation limit.

We then train on the collected states.
For each state, the current policy samples a group of candidate next actions, which receive point-wise rewards and are optimized using GRPO.
These candidates are newly sampled responses, rather than the responses previously used to construct the buffer.
All candidates in a group are evaluated against the same prefix and remaining-reference set; matching one candidate does not remove references for its siblings.
After updating the policy on the buffer, we construct a new buffer using the updated policy.
Thus, RL uses model-generated rather than reference-written prefixes, with periodic refreshes of the training-state distribution.

\subsection{Point-Wise Rewards}

We reuse the soundness, substantiveness, and reference-matching criteria from the one-shot setup.
While unmatched high-quality references remain, a non-redundant review point receives
\[
\begin{gathered}
R(s_t,a_t)=\\[-2pt]
\begin{cases}
-1, & \text{unsound},\\[2pt]
-0.25, & \text{sound but non-substantive},\\[2pt]
+0.1, & \begin{tabular}[t]{@{}l@{}}sound and substantive,\\but unmatched,\end{tabular}\\[2pt]
+1, & \begin{tabular}[t]{@{}l@{}}sound and substantive, matching\\an uncovered reference.\end{tabular}
\end{cases}
\end{gathered}
\]

Points that overlap with the existing prefix receive $-1$.
A reference match requires high relatedness and equal or greater specificity relative to the reference.
Previously covered references cannot earn another matching bonus within the same trajectory.

We also explicitly reward termination.
Let $c_t$ denote the fraction of high-quality references covered by the prefix.
Producing \texttt{<STOP>} receives $+1$ when all such references have been covered, and
\[
-0.5(1-c_t)
\]
otherwise.
Once no high-quality references remain uncovered, further nonterminal generation receives $-0.25$.
These rules encourage additional useful coverage while discouraging repetition and unnecessary continuation.

\subsection{Results}

The result is summarized in Table~\ref{tab:oneshot-vs-recursive}.
The recursive generation model has degraded performance compared to one-shot generation model. SFT results are similar, but the difference occurs at the RL stage.
We conclude that giving review-level F1 reward is sufficient for dense reward modeling.

\begin{table}[h]
\centering
\small
\setlength{\tabcolsep}{3pt}
\renewcommand{\arraystretch}{1.05}

\begin{tabularx}{\columnwidth}{@{}Yccc@{}}
\toprule
\textbf{} & \textbf{Precision} & \textbf{Recall} & \textbf{F1} \\
\midrule

\multicolumn{4}{l}{\textit{Recursive generation}} \\
SFT                     & 18.5 & 30.6 & 17.4 \\
~~+RL w/ point-wise reward  & 42.3 & 42.8 & 37.1 \\
\midrule

\rowcolor{LightGreen} \multicolumn{4}{l}{\textit{One-shot generation (Ours)}} \\
SFT              & 23.0 & 25.6 & 16.6 \\
~~+RL w/ F1 reward  & \textbf{47.4} & \textbf{51.4} & \textbf{45.1} \\

\bottomrule
\end{tabularx}
\caption{
\textbf{Comparison of models trained for one-shot review generation vs. recursive review points generation.}
For RL training under recursive generation, we introduce a review point-wise reward, which gives graded scores based on soundness, substantiveness, and reference-matching.
We evaluate those models on the \texttt{PaperSound-23K} test set.
}
\label{tab:oneshot-vs-recursive}
\end{table}

\section{Prompts}
\label{app:sec:prompts}

We report the instruction text used for \texttt{PaperSound-23K} dataset construction and PaperCritic training.

\subsection{Dataset Construction}
\label{app:subsec:dataset-prompts}

\subsubsection{Extracting human-written review points}
\label{app:prompt-human-extraction}

\noindent\textbf{System.}
\begin{Prompt}
You extract structured review points from ML peer reviews.

Your goal is to mimic a careful human annotator who splits reviewer text into self-contained weakness units while preserving original wording.

Definition:
- A review point is one semantically independent weakness or critical question.
- It must stand on its own.
- Preserve reviewer wording as directly as possible.

Segmentation rules:
1) Start from bullets, numbering, indentation, and paragraph breaks.
2) Merge fragments when they are one claim plus its explanation/evidence/follow-up ask.
3) Split only when they are clearly different claims with different rationale.
4) If a parent bullet/heading is needed to identify the target of a child bullet, include BOTH in the extracted point.
5) If a parent line is only context/topic and multiple sibling child bullets each contain their own standalone ask, output multiple points and repeat the parent context in each one.
6) Within a single child bullet, numbered item, or paragraph, keep all same-target follow-up sentences/questions together when they elaborate the same concern.
7) For long paragraphs with an opening sentence, enumerated premises, and a concluding concern, keep the FULL same-claim block together, including the enumerated middle.
8) If the criticism depends on the reviewer's own understanding or intuition, keep that context in the same point. If later text says "Given this intuition", "To me, the reason ...", "From my understanding ...", or refers back to (i)/(ii)/(iii), include the preceding explanatory context block.
9) If a later sentence explicitly shifts to a new request with wording like "A minor suggestion would be ...", split it when that later sentence can stand alone as a different request from the preceding limitation.
10) Remove duplicate recap points that merely summarize earlier concrete criticisms.

Drop these:
11) Strengths/praise.
12) Pure cross-references or navigation text with no standalone criticism.
13) Opening review-summary sentences or closing verdict/rebuttal paragraphs that only summarize the reviewer's overall stance rather than stating a concrete standalone weakness.
14) Broad verdict-only statements that do not add a concrete issue beyond nearby detailed points.

Text fidelity:
15) Extract spans from reviewer text; do not paraphrase, summarize, simplify, or generalize.
16) Keep citations exactly as written if the reviewer wrote them.
17) Do NOT remove citations, author-year mentions, URLs, paper titles, or reference markers from weakness text.
18) Do NOT rewrite a point just to clean citations. Preserve the original wording instead.
19) Minimal local rewriting is allowed only to repair an otherwise broken pronoun/deictic reference needed for standalone meaning.
20) If a reviewer or author appends standalone reference lines immediately below a question or comment, keep those reference lines attached to the same extracted weakness/response span.

References field:
21) references.reviewer and references.authors should contain only literal citation strings that appear verbatim in the extracted reviewer weakness text or author response text, such as [1], [4,5], "X et al. (2020)", "X et al., 2020", bare URLs, markdown links, or arXiv IDs.
22) Do NOT include model names, dataset names, section/figure/equation mentions, paper-topic phrases, inferred author names, shortened partial forms, or prose summaries in references.*.

Response mapping:
22) Map each weakness to the best author response span copied verbatim.
23) No author-response paraphrase, no ellipsis, no truncation.
24) If there is no response, output "None".

Examples:
25) If the source is a parent context line plus multiple sibling child bullets that each ask different questions, output multiple points and repeat the parent context in each point.
26) If the source is a single numbered concern followed by multiple short questions about the SAME section/object, keep them as ONE point.
27) If the source is a long paragraph with an opening sentence, an (i)/(ii)/(iii) block, and a concluding concern, keep the full block as ONE point.
28) If later text says "Given this intuition" or "From my understanding", include the preceding explanatory context in the same point.
29) If the source is an opening high-level summary sentence of the review or a closing rebuttal/score paragraph, drop it.
30) If the source is "See my comments on clarity.", drop it.
31) If the source is "Some baselines such as [1] are not considered and should be added.", keep it exactly, including "[1]".
32) If the source is "Kveton et al. 2019 - double citation.", keep it exactly.
33) If the source contains "Kumar et al. 2021", "Ho et al.", or a URL, keep that citation text if it is present in the reviewer comment.

Return JSON only matching schema.
\end{Prompt}

\noindent\textbf{User.}
\begin{Prompt}
Paper title: <TITLE>
Year: <YEAR>
Reviewer ID: <REVIEWER_ID>

[REVIEWER TEXT]
<REVIEWER_TEXT>

[AUTHOR RESPONSE THREAD FOR THIS REVIEW]
<AUTHOR_RESPONSE_THREAD>

[GLOBAL AUTHOR RESPONSE (OPTIONAL CONTEXT)]
<GLOBAL_AUTHOR_RESPONSE>
\end{Prompt}

\noindent\textbf{Structured response contract (schema summary, not additional prompt text).}
\begin{Prompt}
{"weakness_response_pairs": [{"weakness": "<VERBATIM_WEAKNESS>", "response": "<VERBATIM_RESPONSE_OR_None>"}],
 "references": {"reviewer": ["<LITERAL_CITATION>"], "authors": ["<LITERAL_CITATION>"]}}
\end{Prompt}

\subsubsection{Generating model-written review points}
\label{app:prompt-model-candidates}

\noindent\textbf{System.}
\begin{Prompt}
You are an expert anonymous reviewer for ICLR 2020-2023 submissions.

Generate only reviewer weakness points as schema-constrained structured output.
Each output item must be a standalone, paper-specific weakness or substantive reviewer question.
Do not output a summary, strengths, scores, recommendation, markdown headings, or any free-form text outside the schema.
\end{Prompt}

\noindent\textbf{User.}
\begin{Prompt}
ICLR 2020-2023 review context:
- Reviewers assess soundness, presentation, contribution, strengths, weaknesses, questions for the authors, ethics concerns, and an overall rating.
- For dataset construction, return only the weaknesses/questions content as independent weakness units.

Task:
- Read the paper and identify the distinct concerns that a strong ICLR reviewer would raise before author rebuttal.
- Focus on concerns that matter to the paper's main contributions.
- Prefer paper-grounded, specific critiques over generic checklist comments.
- Use reviewer questions only when the question exposes a substantive uncertainty or missing piece of evidence.

Inclusion priorities:
1. Unsupported, overstated, or weakly evidenced central claims.
2. Missing, weak, unfair, or insufficient empirical validation.
3. Theoretical gaps, unjustified assumptions, unclear derivations, or unproven steps.
4. Novelty or positioning issues that materially affect the claimed contribution.
5. Reproducibility, limitations, or ethics issues that materially affect the work.
6. Clarity problems that obstruct correct evaluation of the main claims.

Exclusion rules:
1. No strengths, summary, rating, confidence, or accept/reject language.
2. No trivial copyediting or formatting complaints unless they create technical ambiguity.
3. No duplicate points or near-duplicates.
4. Do not split one concern into multiple fragments.
5. Do not bundle unrelated concerns into one item.
6. Do not invent missing baselines, related work, or experiments unless they are clearly implied by the paper's own claims and setup.
7. Do not rely on post-2023 hindsight or external facts not available from the submission itself.

Granularity guidance:
- Bad: "The experiments are weak."
- Good: "The empirical support for the main claim is incomplete because the paper omits comparisons against strong baselines on the hardest evaluation setting, so it is unclear whether the reported gains come from the proposed method or the evaluation setup."
- Bad: "Clarity is poor."
- Good: "Section 3 does not define the optimization objective precisely; the notation changes between the main equation and the algorithm box, making it difficult to verify what is actually optimized."

Style guidance:
- Each `review_point` should stand alone semantically.
- Each `review_point` should be concise but specific, usually 1-3 sentences.
- It is fine to refer to sections, equations, figures, or tables when that sharpens the criticism.
- Usually return 5-9 non-redundant points, but do not pad with weak comments.
- Return an empty list only if there are truly no substantive weaknesses, which should be rare.

Structured output reminder:
- Populate only `review_points`.
- Each element must be an object with one field: `review_point`.

Paper title: <TITLE>
Conference year: <YEAR>

Paper:
<PAPER_TEXT>
\end{Prompt}

\noindent\textbf{Structured response contract (schema summary, not additional prompt text).}
\begin{Prompt}
{"review_points": [{"review_point": "<CANDIDATE_REVIEW_POINT>"}]}
\end{Prompt}

\subsubsection{Extracting main claims and evidence}
\label{app:prompt-claims}

\noindent\textbf{System.}
\begin{Prompt}
Follow the user's claim-extraction instructions exactly. Return only JSON matching the schema.
\end{Prompt}

\noindent\textbf{User.}
\begin{Prompt}
You extract only the decision-relevant main claims of ML papers and the concrete paper-internal evidence supporting each claim.
Output JSON only and make it satisfy the provided schema exactly.

Definition of a main claim:
- A central contribution or headline finding that should matter to a review decision.
- It should be one of the paper's strongest takeaways from the abstract, introduction, theorem/results sections, or major empirical results.

Selection and coverage rules:
- Return 3-6 claims ordered by importance, but prefer 3-5.
- It is better to return 4 strong claims than 5 claims with one weak or ancillary point.
- If the paper introduces a new method, objective, algorithm, model family, measure, or analysis framework, one of the first 1-3 claims must explicitly describe that technical contribution itself.
- If the paper also has major theory or major empirical findings, include them as separate claims; do not let performance claims crowd out the main technical contribution.
- Exclude minor ablations, robustness checks, diagnostics, prompt variations, side observations, implementation tricks, feature choices, hyperparameter observations, and engineering details unless the paper clearly presents them as headline contributions.
- Exclude background, motivation, positioning against prior work, future work, and generic framing.

Claim-writing rules:
- One claim = one self-contained assertion.
- Keep the claim specific but not overloaded.
- Do not merge multiple independent findings into one claim.
- In particular, do not combine separate result families such as data efficiency and robustness, or method definition and a secondary side observation, into one claim.
- Be faithful and conservative.

Evidence rules:
- For each claim, provide 1-4 evidence bullets.
- Every evidence bullet must point to concrete support inside the paper and should explicitly mention at least one of these when possible: theorem/proposition/lemma/corollary, equation/definition, algorithm, table, figure, experiment, or a quantitative result.
- Never use the abstract, contribution list, introduction summary, or conclusion summary as an evidence bullet by themselves. Those sections are useful for judging centrality, not for evidential support.
- Generic motivation, marketing prose, and prior-work contrast do not count as evidence.
- Intuition or interpretation is allowed only if it is tied to a concrete theorem/equation/figure/table/experiment in the same bullet.
- Do not simply restate the claim in different words.
- If the claim is empirical, prefer quantitative results.
- If the claim is theoretical, prefer the exact theorem/proposition/condition.
- If the claim is about a new method validated empirically, prefer evidence that includes both the method definition and its main validating result.

Paper title: <TITLE>
Year: <YEAR>

Paper content:
<PAPER_TEXT>
\end{Prompt}

\noindent\textbf{Structured response contract (schema summary, not additional prompt text).}
\begin{Prompt}
{"claims": [{"id": "<CLAIM_ID>", "claim": "<CLAIM_TEXT>",
             "supporting_evidence": ["<EVIDENCE_BULLET>"]}]}
There are 3-6 claims and 1-4 supporting-evidence bullets per claim.
\end{Prompt}

\subsubsection{Screening for paper-soundness relevance}
\label{app:prompt-screening}

\noindent\textbf{System.}
\begin{Prompt}
You screen one peer-review point for whether it evaluates paper soundness. Return only JSON that matches the schema.
\end{Prompt}

\noindent\textbf{User.}
\begin{Prompt}
Paper title: <TITLE>

Paper full text:
<PAPER_TEXT>

Main claims with supporting evidence:
<CLAIMS_WITH_EVIDENCE>

Review point:
<REVIEW_POINT>

Task:
Decide whether this review point evaluates the paper's soundness.

Use an inclusive evidence-screen, not a narrow 'formal theorem only' screen.
Mark evaluates_paper_soundness=true whenever the point argues that the paper has not yet sufficiently established, distinguished, or justified at least one main claim or core contribution. This includes missing evidence, but also central complaints that the claimed contribution is not clearly distinct, not clearly defined, not clearly necessary, or not clearly the right interpretation of what the paper shows.

This includes:
- missing baselines or unfair comparisons that weaken a performance claim
- missing ablations, datasets, or experimental settings that weaken a generality claim
- unclear optimization/training/evaluation details when that blocks judging a claim
- overclaiming, unsupported extrapolation, weak theoretical assumptions, or missing proof steps
- requests for additional evidence when the current evidence may be insufficient for a main claim
- close-related-work / novelty / positioning critiques when they argue the paper has not established what is genuinely new or distinct about a central contribution
- 'I cannot disentangle what the contributions are' when that means the core claim is not stated or supported clearly enough to evaluate
- objections that a named construct may not actually fit the claimed concept, terminology, or interpretation
- complaints that the paper's technical depth or support is too limited for the central contribution it is asking readers to accept

Mark evaluates_paper_soundness=false only when the point is purely about wording, organization, jargon, formatting, bibliography completeness, generic future work, or novelty/positioning comments that do not materially challenge whether a main claim or core contribution is established, distinct, or justified.

Important edge cases:
- 'Please add baseline X because otherwise the superiority claim is not convincing.' -> true
- 'Please cite related paper Y.' -> false unless it also says the paper has not established what is new/distinct, or that a central claim is not justified without that comparison.
- 'Please explain optimization details because I cannot tell how the reported result was obtained.' -> true
- 'I cannot tell what the paper's actual contribution is or how it differs from close prior work.' -> usually true
- 'If this process is non-Markovian/non-continuous, does it still make sense to call it diffusion?' -> true when this is really questioning whether the paper's claimed object or interpretation is correct
- 'The wording of Definition 2 is awkward.' -> false
- 'This is not new' with no explanation of why that undermines the paper's claimed contribution -> false

If the point is mixed, choose true whenever the evidence-gap component is substantive.
Return strict JSON only.
\end{Prompt}

\noindent\textbf{Structured response contract (schema summary, not additional prompt text).}
\begin{Prompt}
{"evaluates_paper_soundness": <BOOLEAN>, "filter_reason": "<REASON>",
 "screening_rationale": "<RATIONALE>"}
REASON is one of: soundness_related, novelty_or_significance_only,
presentation_or_clarity_only, related_work_or_citation_only,
administrative_or_ethics_only, other_non_soundness.
\end{Prompt}

\subsubsection{Routing a review point to a main claim}
\label{app:prompt-dataset-routing}

\noindent\textbf{System.}
\begin{Prompt}
You route one soundness-relevant review point to the single paper claim it most directly challenges. Return only JSON that matches the schema.
\end{Prompt}

\noindent\textbf{User.}
\begin{Prompt}
Paper title: <TITLE>

Paper full text:
<PAPER_TEXT>

Main claims with supporting evidence:
<CLAIMS_WITH_EVIDENCE>

Review point:
<REVIEW_POINT>

Task:
Choose the single main claim most directly challenged by this review point.

Routing rules:
- Pick the claim whose truth would be most weakened if the review point were accepted.
- Do not mechanically prefer the narrowest downstream result claim. Many review points primarily attack the paper's main contribution/method claim even if a specific experiment or theorem is the immediate evidence gap.
- If the point questions novelty, distinctness from related work, what the contribution actually is, whether the method/framework is generally useful, whether a core component is well-defined, or whether the approach only works in a narrower setting than claimed, usually route to the primary contribution/method claim.
- If the point asks for a missing baseline, comparison, ablation, dataset, hyperparameter study, or user study, route to the claim the reviewer is trying to validate. Use the broader method/contribution claim when the reviewer is asking whether the overall method is useful/necessary/established, and use a narrower empirical claim only when the point is clearly about one specific reported result.
- If the point questions optimizer mismatch, theorem assumptions, proof details, or theory-vs-experiment consistency, route to the corresponding theoretical or theory-backed claim.
- If the point challenges whether an observed pattern really supports the paper's proposed mechanism or interpretation, route to the mechanism/theory-link claim rather than the raw observation claim.
- If the point asks about runtime, pruning overhead, computational savings, parameter sharing, or a core mapping/component needed for the method to work as described, route to the method/practicality/theory claim that depends on that detail, not automatically to the final benchmark claim.
- If several claims are affected, still pick one primary target and up to two secondary candidates for diagnostics.
- Use null only if no listed claim can be linked in a defensible way.

Important edge cases:
- 'This seems close to prompt tuning / prior work X / not very new.' -> usually route to the primary contribution claim.
- 'I cannot tell what the contribution is' or 'I cannot disentangle which components are necessary.' -> usually route to the primary contribution claim.
- 'This only seems to work for text-based games / tokenizable states.' -> usually route to the primary contribution or scope claim, not a later benchmark result.
- Missing user study or external validation should route to the broader contribution/usefulness claim unless a listed claim explicitly centers on user alignment or human evaluation.
- Missing optimizer or training-detail concerns should route to the claim whose justification depends on that setup.
- Missing dataset/ablation/baseline concerns should route to the broader method claim when the reviewer is asking whether the method is established at all, and to a narrower empirical claim only when the point clearly targets that specific result.
- 'Dot products may rise because norms grow, so the claimed mechanism may not follow.' -> route to the mechanism/theory-link claim, not just the empirical observation claim.
Return strict JSON only.
\end{Prompt}

\noindent\textbf{Structured response contract (schema summary, not additional prompt text).}
\begin{Prompt}
{"target_claim_id": "<CLAIM_ID_OR_NULL>",
 "secondary_target_claim_ids": ["<CLAIM_ID>"], "routing_rationale": "<RATIONALE>"}
Use JSON null, not a string, when no primary claim applies.
The secondary-claim list contains at most two identifiers.
\end{Prompt}

\subsubsection{Annotating review-point soundness}
\label{app:prompt-dataset-soundness}
The three premise verdicts correspond to soundness scores 0, 1, and 2, respectively, in the order listed in the decision rule.

\noindent\textbf{System.}
\begin{Prompt}
You verify whether a review point's premises are actually supported by the paper. Be skeptical, identify material premises explicitly, and use the paper as the source of truth. Return only JSON that matches the schema.
\end{Prompt}

\noindent\textbf{User.}
\begin{Prompt}
Paper title: <TITLE>

Paper full text:
<PAPER_TEXT>

Main claims with supporting evidence:
<CLAIMS_WITH_EVIDENCE>

Proposed target claim: <TARGET_CLAIM_ID>

Review point:
<REVIEW_POINT>

Task:
Judge whether the material premises of the review point are actually correct.

First identify the material premises: the factual assumptions that must be true for the criticism to hold.

For each material premise, use this evidence order:
1. Check whether the paper itself supports, contradicts, or leaves the premise unresolved.
2. If the paper alone does not settle the premise, you may use your internal knowledge.
Treat the paper's publication/reveal year as <YEAR>. If you need internal knowledge beyond the paper text, only use knowledge that plausibly predates or is contemporaneous with <YEAR>; do not rely on later developments.
3. If the premise is still not clearly judgeable after using both the paper and eligible prior knowledge, treat it as unresolved.

Use this decision rule exactly:
- Choose premise_verdict=contains_material_false_premise if at least one material premise is clearly false, whether judged from the paper or from eligible prior knowledge. This includes paper contradictions, reviewer misunderstandings, garbling what the paper says, logically invalid core inferences, and false claims about prior work or standard facts.
- Choose premise_verdict=not_fully_verifiable_but_no_material_false_premise only if no material premise is clearly false, but at least one material premise still cannot be clearly judged after checking both the paper and eligible prior knowledge. This includes unresolved value-laden or underspecified premises.
- Choose premise_verdict=fully_supported_by_paper_or_missing_evidence_confirmed only if every material premise is true, whether judged from the paper or from eligible prior knowledge. A premise can also count as true when the review point says the paper lacks certain evidence/detail and the paper indeed lacks it.

Critical boundary between 0 and 1:
- There is no 'mixed but still 1' bucket. If a point contains one valid concern and one material false premise, choose contains_material_false_premise.
- Do not average across clauses. Score by the weakest material factual premise.
- Polite question wording does not rescue a false embedded premise.
- If the point says or implies 'no evidence', 'never explained', 'statistical significance is not there', or proposes a concrete missing setup, and the paper already provides some relevant evidence/explanation/setup, that is 0, not 1.
- If the paper already reports uncertainty evidence that materially addresses the point (confidence intervals, standard errors, error bars, bootstrap CIs, repeated-seed variability), then claiming significance evidence is absent is usually 0, not 1.
- If the point cites undefined comparisons, the wrong equation/object, or a garbled formal quantity, and that mistake is material to the criticism, that is 0.
- If the point starts from a correct observation but then leaps to a categorical conclusion such as 'therefore this implies nothing' or 'therefore the method is unmotivated,' and that leap does not follow from the paper, that is 0.
- If the point asks for better positioning, related work, or clarification without embedding a false factual claim, that can be 1.
- If the paper truly lacks the named experiment, baseline, ablation, proof step, compute comparison, or explanation, that can be 2.

Also detect whether the review point itself explicitly cites or relies on external work. Mark cites_external_work_detected_by_llm=true only when the review point itself explicitly references external work, for example by using author-year citations, numbered citations like [1], paper titles, URLs, arXiv identifiers, named prior methods/papers used as references, or similar literal citation cues. Do not mark true just because the paper or authors' response mentions prior work. Set external_citation_evidence to a short literal snippet from the review point, or 'none' if nothing is detected.

Representative examples:
- 'How were GPU hours computed, from fine-tuning or pre-training+fine-tuning?' -> 0 if the paper already states the table is pre-training-only, even if other accounting details remain unclear.
- 'This is never explained.' -> 0 if the paper does give an explanation, even if the explanation is brief or unconvincing.
- 'The paper provides no empirical evidence.' -> 0 if experiments relevant to that point are present.
- 'Statistical significance is not there.' -> 0 if the paper already reports uncertainty intervals or repeated-seed variability that materially addresses this concern.
- 'Equation/object X behaves this way' -> 0 if the paper does not define X that way, or the point confuses different equations/conditionals.
- 'Because quantity Y changes with the iterate, the paper's interpretation implies nothing.' -> 0 if the observation about Y is true but the categorical conclusion does not follow.
- 'Could authors position this relative to prompt tuning / related work X?' -> 1 if this is a plausible request and does not rely on a false factual claim.
- 'The paper does not report runtime / compute comparisons for the claimed efficiency gain.' -> 2 if such evidence is indeed missing.

Return:
- premise_verdict
- false_premise_type: contradicted_by_paper, reviewer_overstatement, reviewer_misunderstanding, or none
- core_premises_summary: concise summary of the material premises you checked
- soundness_rationale: explain why the chosen verdict follows from the paper and, if needed, eligible prior knowledge
- cites_external_work_detected_by_llm: boolean
- external_citation_evidence: short literal snippet from the review point, or 'none'
Do not return a numeric score; it will be derived from premise_verdict.
\end{Prompt}

\subsubsection{Annotating claim linkage and severity}
\label{app:prompt-dataset-substantiveness}

\noindent\textbf{System.}
\begin{Prompt}
You score claim linkage and severity for one review point against a selected target claim. Return only JSON that matches the schema.
\end{Prompt}

\noindent\textbf{User.}
\begin{Prompt}
Paper title: <TITLE>

Paper full text:
<PAPER_TEXT>

Main claims with supporting evidence:
<CLAIMS_WITH_EVIDENCE>

Selected target claim: <TARGET_CLAIM_ID>
Claim-routing rationale: <ROUTING_RATIONALE>
Prior soundness assessment: soundness=<SOUNDNESS_SCORE>, premise_verdict=<PREMISE_VERDICT>

Review point:
<REVIEW_POINT>

Task:
Score claim_linkage and severity for the selected target claim.

Use these criteria exactly:

B. Claim Linkage
"How directly does this review point target the paper's main claim?"
- 0: Unrelated - Irrelevant to the core claim (e.g., formatting issues, minor details, or unrelated topics).
- 1: Peripheral - Addresses peripheral aspects (e.g., auxiliary experiments, presentation choices, minor assumptions).
- 2: Adjacent - Targets an important premise or supporting evidence closely related to the main claim.
- 3: Direct - Directly challenges the validity of the main claim itself.

C. Severity
"Assuming the review point is sound, how damaging is it to the paper's main claim?"
- 0: Cosmetic - Minor issues (e.g., wording, organization, small mistakes) with little impact on the conclusion.
- 1: Weak Evidence Gap - Weakens the evidence somewhat, but the issue could likely be fixed with additional support.
- 2: Moderate Risk - Creates a moderate to significant risk that the central evidence may not hold.
- 3: Fatal Flaw - Breaks a key line of reasoning; the main conclusion is unlikely to hold.

Important scoring guidance:
- Score linkage relative to the selected target claim, not the entire paper.
- Do not re-route to a different claim unless the selected claim is clearly impossible; if impossible, assign claim_linkage=0 and explain briefly.
- Requests for more experiments/baselines can still have linkage 2 or 3 if those experiments are necessary to support the selected claim.
- Do not inflate severity for comments that are mainly editorial, presentational, or only weakly connected to the selected claim.
- Severity is conditional on the point being sound; use the prior soundness result as context, but still imagine the point's premise is true when assigning severity.
Return strict JSON only.
\end{Prompt}

\noindent\textbf{Structured response contract (schema summary, not additional prompt text).}
\begin{Prompt}
{"claim_linkage": <INTEGER_0_TO_3>, "severity": <INTEGER_0_TO_3>,
 "scoring_rationale": "<RATIONALE>"}
\end{Prompt}

\subsection{PaperCritic Training}
\label{app:subsec:training-prompts}

\subsubsection{Policy prompt}
\label{app:prompt-oneshot-policy}

No separate system message is supplied.

\noindent\textbf{User.}
\begin{Prompt}
# Paper
<PAPER_TEXT>

# Instruction
Write a list of high-quality review points for the paper.
Requirements:
- Each review point must assess the soundness of the paper: whether the paper's claims are supported by accurate, convincing, and clear evidence.
- Each review point must contain correct premises and avoid relying on false premises.
- Each review point must create a moderate to significant risk that a main claim or its central evidence may not hold.
- Do not include review points about novelty, writing quality, missing citations, or minor presentation issues unless they directly undermine the soundness of a main claim.
- Output only the review points as a numbered list, with one review point per item.
\end{Prompt}

\subsubsection{Target-claim routing}
\label{app:prompt-rl-routing}

\noindent\textbf{System.}
\begin{Prompt}
You are a strict reviewer-evaluation judge. Return only valid JSON.
\end{Prompt}

\noindent\textbf{User.}
\begin{Prompt}
# Paper
<PAPER_TEXT>

# Main Claims and Evidence
<CLAIMS_WITH_EVIDENCE>

# Generated Review Point
<GENERATED_POINT>

# Instruction
Identify the most relevant target main claim of the generated review point.
Requirements:
- Output only valid JSON with exactly one key: target_claim_id.
- "target_claim_id" must be "Cx".
- Do not include any rationale or additional text.
\end{Prompt}

\subsubsection{Soundness reward model prompt}
\label{app:prompt-soundness-rm}

The system message is shared with Section~\ref{app:prompt-rl-routing}.

\noindent\textbf{User.}
\begin{Prompt}
# Paper Title
<TITLE>

# Paper
<PAPER_TEXT>

# Target Review Point
<REVIEW_POINT>

# Instruction
Task:
Judge whether the material premises of the review point are actually correct.

First identify the material premises: the factual assumptions that must be true for the criticism to hold.

For each material premise, use this evidence order:
1. Check whether the paper itself supports, contradicts, or leaves the premise unresolved.
2. If the paper alone does not settle the premise, you may use your internal knowledge.
Treat the paper's publication/reveal year as <YEAR>. If you need internal knowledge beyond the paper text, only use knowledge that plausibly predates or is contemporaneous with <YEAR>; do not rely on later developments.
3. If the premise is still not clearly judgeable after using both the paper and eligible prior knowledge, treat it as unresolved.

Use this decision rule exactly:
- Choose score=0 if at least one material premise is clearly false.
- Choose score=1 only if no material premise is clearly false, but at least one material premise still cannot be clearly judged.
- Choose score=2 only if every material premise is true. A premise can also count as true when the review point says the paper lacks certain evidence/detail and the paper indeed lacks it.

Critical boundary between 0 and 1: there is no mixed-but-still-1 bucket. If a point contains one valid concern and one material false premise, choose score=0. Do not average across clauses.

Return strict JSON only with exactly these keys:
- score: one of 0, 1, 2
- rationale: an object with premise_reconstruction and premise_factuality_judgment
\end{Prompt}

\subsubsection{Substantiveness reward model prompt}
\label{app:prompt-substantiveness-rm}

The system message is shared with Section~\ref{app:prompt-rl-routing}.

\noindent\textbf{User.}
\begin{Prompt}
# Paper
<PAPER_TEXT>

# Target Main Claim and Evidences
<TARGET_CLAIM_AND_EVIDENCE>

# Rubric Definitions
## Claim Linkage
0: The concern is not linked to a main claim.
1: The concern is weakly or indirectly linked to a main claim.
2: The concern is clearly linked to a main claim.
3: The concern directly attacks a central main claim or its key evidence.

## Severity
0: The concern has negligible impact.
1: The concern has minor impact.
2: The concern creates a moderate risk to the claim.
3: The concern creates a major risk to the claim.

# Relevant Reference Examples
These examples are provided only to calibrate the rubric for this paper.

<REFERENCE_EXAMPLES_WITH_CLAIM_LINKAGE_AND_SEVERITY>

# Generated Review Point
<GENERATED_POINT>

# Instruction
Evaluate only claim_linkage and severity for the generated review point. Requirements:
- Do not score, discuss, or infer soundness; factual correctness is judged separately by another reward model.
- For severity, assume the review point's factual premises are correct and ask how damaging the concern would be to the selected target claim.
- Output only valid JSON with exactly these two keys: claim_linkage, severity.
- claim_linkage and severity must each be one of {0, 1, 2, 3}.
- Do not include any rationale or additional text.
\end{Prompt}

\subsubsection{Reference-matching prompt}
\label{app:prompt-matcher}

\noindent\textbf{System.}
\begin{Prompt}
You compare one generated review point against candidate reference review points from the same paper. Return only JSON matching the schema.
\end{Prompt}

\noindent\textbf{User.}
\begin{Prompt}
# Paper Title
<TITLE>

# Generated Review Point
<GENERATED_POINT>

# High-Quality Review Points
<REFERENCE_POINT_ID> [<SOURCE>]: <REFERENCE_POINT_TEXT>

# Instruction
For each high-quality review point, judge:
- relatedness: one of {none, weak, medium, high}
- relative_specificity: compare the generated review point to the high-quality review point using one of {less, same, more}

Judging relatedness:
- high: they express essentially the same criticism, or the generated point clearly subsumes the high-quality point at the issue level
- medium: they are about a similar weakness area but not the same concrete criticism
- weak: they only share a broad topic
- none: they are different issues

Judging specificity:
- less: the generated point is broader or vaguer than the high-quality point
- same: roughly the same granularity/specificity
- more: the generated point is narrower or more detailed while still covering the same issue

# Specificity Calibration
Judge semantic issue scope, not writing length or the number of technical details.
- A long generated point can still be less specific when it asks a broad question, covers many possible causes/settings, or omits the reference point's narrow target or condition.
- A short reference point can be highly specific when it identifies one concrete table, equation, hyperparameter, comparison, assumption, or failure case.
- Use same when the two criticisms operate at approximately the same issue granularity, even if one is wordier.
- Use more only when the generated point retains the reference point's core target and conditions and then narrows them further. Extra examples, citations, proposed experiments, or background do not by themselves make it more specific.
- Relatedness and specificity are separate judgments. Even when relatedness is none, weak, or medium, still compare the semantic breadth/granularity of the two criticisms instead of defaulting to same or more.

A high-quality point is matched only if relatedness is high and relative_specificity is same or more.
Return all matched high-quality review points. If no point satisfies the matching rule, return null.

Output JSON only in exactly this shape:
{
    "matches": [
      {
        "hq_point_id": "HQ1",
        "relatedness": "high",
        "relative_specificity": "same",
        "rationale": "Brief reason for matching HQ1."
      },
      {
        "hq_point_id": "HQ2",
        "relatedness": "high",
        "relative_specificity": "more",
        "rationale": "Brief reason for matching HQ2."
      }
    ]
  }
\end{Prompt}

\end{document}